\documentclass[10pt,twocolumn,letterpaper]{article}

\usepackage[pagenumbers]{cvpr} 

\usepackage{pifont}           
\usepackage{fontawesome5}     
\usepackage{float}            
\usepackage{listings}         
\usepackage[most]{tcolorbox}  
\usepackage{xr-hyper}         

\usepackage{microtype}

\definecolor{cvprblue}{rgb}{0.21,0.49,0.74}
\usepackage[pagebackref,breaklinks,colorlinks,allcolors=cvprblue]{hyperref}

\definecolor{PromptFrame}{RGB}{92, 112, 128}   
\definecolor{PromptBack}{RGB}{248, 250, 252}   
\definecolor{dynamicblue}{RGB}{36, 99, 158}
\newcommand{\dyn}[1]{\textcolor{dynamicblue}{[#1]}}

\def\paperID{144} 
\def\confName{3DV\xspace}
\def\confYear{2027\xspace}

\title{$\phi$-Scene: Physically Grounded Image-to-3D Scene Reconstruction}

\author{
Haodong Li, Lulu Shao, Haolin Lu, Yu Fu, Yen-Ru Chen, Seemandhar Jain, Manmohan Chandraker\\
University of California San Diego\\
\url{https://phi-scene.github.io/}
}

\begin{document}

\twocolumn[{%
    \renewcommand\twocolumn[1][]{#1}%
    \maketitle
\vspace{-.6cm}
\begin{center}
    \captionsetup{type=figure}
    \includegraphics[width=\textwidth]{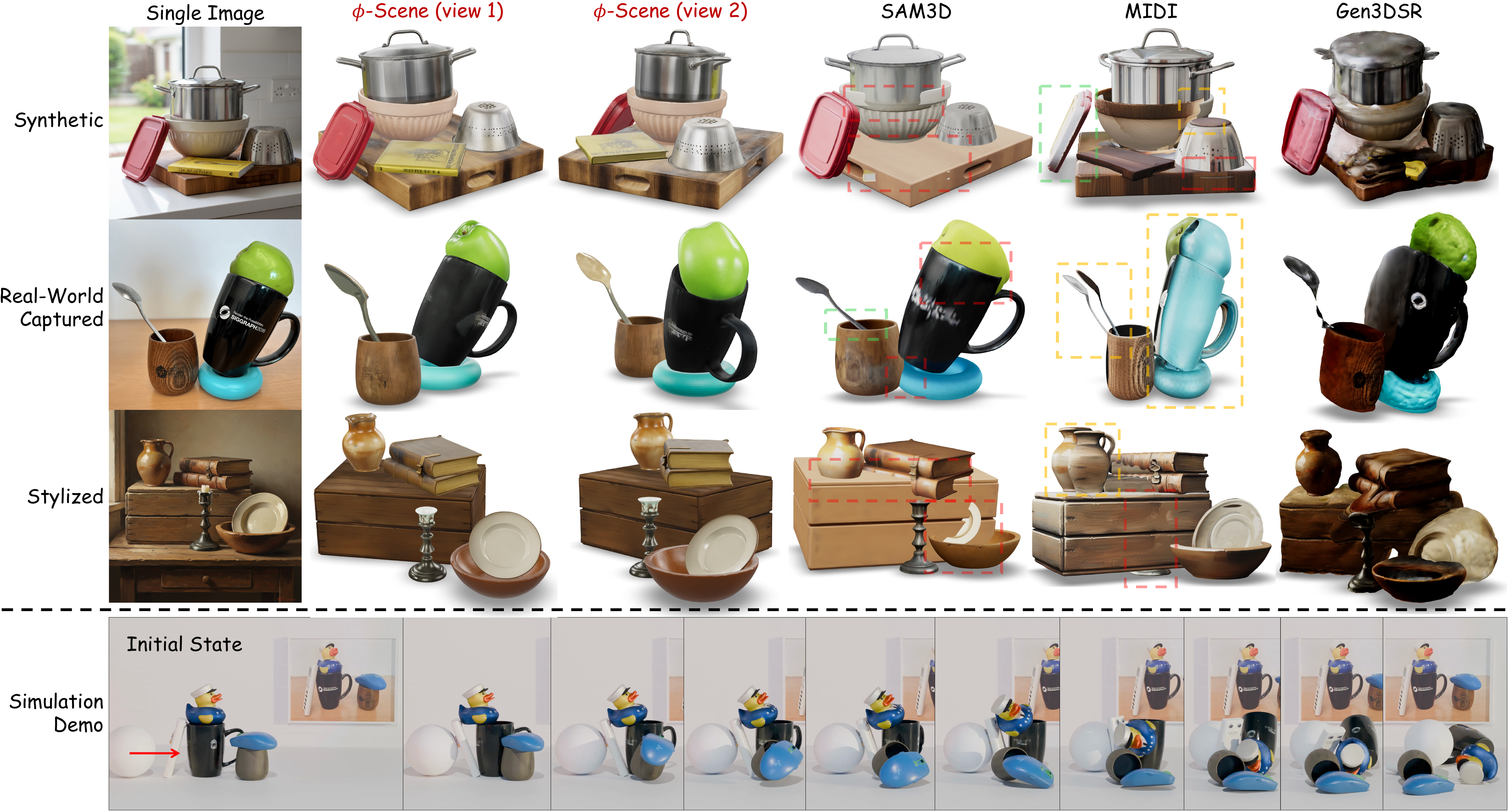}
\vspace{-.6cm}
\caption{
\textbf{From a single image, $\phi$-Scene reconstructs a 3D scene that is both reference aligned and physically stable.}
$\phi$-Scene infers how objects support one another and settles each in topological order against its already-stabilized context.
The resulting 3D scene stays reference-aligned, explicitly reaches dynamic rigid-body equilibrium, and is directly simulation-ready (bottom row).
To our knowledge, $\phi$-Scene is among the first image-to-3D scene reconstruction methods that explicitly reaches dynamic rigid-body equilibrium while preserving reference alignment.
Penetrations, floatings, and irregular objects are boxed in red, green, and yellow, respectively.
}
\vspace{.2cm}
  \label{fig:teaser}
\end{center}
}]

\begin{abstract}
Recent image-to-3D scene methods recover high-fidelity 3D objects with plausible arrangements, but often leave floatings and interpenetrations that limit physical validity and downstream use in interactive environments.
We present $\phi$-Scene, a physically grounded approach for open-vocabulary and compositional image-to-3D scene reconstruction that treats a scene not merely as a set of objects with predicted poses, but as a globally stable physical system.
$\phi$-Scene formulates reconstruction as topology-driven physical assembly: given an initial scene of reconstructed objects, it infers how objects support one another and settles them one by one in topological order.
For each object, SDF-based optimization first resolves penetrations against the already-settled support context, and rigid-body simulation then settles the object into a stable equilibrium under real-world physical constraints.
The resulting scene stays aligned to the reference image, with every object resting at a physically valid, stable contact configuration.
On the 3D-Front benchmark, $\phi$-Scene achieves the strongest overall performance among out-of-domain methods and remains highly competitive with in-domain baselines on standard reconstruction metrics.
Human and MLLM studies prefer $\phi$-Scene in visual quality, reference alignment, and physical plausibility.
Dedicated physical metrics show that it substantially reduces penetration artifacts and yields much lower post-simulation drift.
To our knowledge, $\phi$-Scene is among the first image-to-3D scene reconstruction methods that explicitly reaches dynamic rigid-body equilibrium while preserving reference alignment.
\end{abstract}

\section{Introduction}
\label{sec:intro}

Reconstructing compositional 3D scenes from single images enables digital twins of real-world environments for simulation, robotics, and 3D gaming.
A practically useful image-to-3D scene reconstruction method should produce 3D scenes that are not only coherent with the reference image, but also physically grounded, i.e., with objects arranged into physically stable configurations under real-world physical constraints.

A physically grounded 3D scene should satisfy both static contact validity and dynamic stability.
Static contact validity means that contacting objects exhibit minimal geometric penetration and floating.
Dynamic stability means that the reconstructed scene already reaches a stable rigid-body equilibrium: when released and simulated under physical constraints, objects undergo only small post-simulation drift.

Recent methods have made substantial progress in jointly recovering high-quality individual objects and predicting plausible scene arrangements~\cite{ardelean2025gen3dsr,yao2025cast,meng2025scenegen,zhou2024zero,zhao2025depr,chen2025sam,shi2025scenemaker}.
However, most still treat scene reconstruction primarily as a geometric prediction problem: they estimate objects and poses but do not make physical validity a first-class objective, so reconstructed scenes may contain floating objects, interpenetrations, or other unstable contacts, especially in contact-rich scenes where each object's feasible pose is tightly coupled to its neighbors.

Recent physically oriented methods begin to address this, but each has a key limitation.
CAST~\cite{yao2025cast} treats physical plausibility as a static geometric constraint and resolves penetrations through a joint SDF-based optimization over all object poses without modeling chained support dependencies, so the optimization keeps trading one violation for another instead of reaching a globally valid configuration.
PAT3D~\cite{lin2025pat3d}, in contrast, adds rigid-body simulation but targets text-to-3D generation rather than image-to-3D reconstruction, which requires alignment to a given reference image; it further relies on a top-down single-parent support tree that cannot handle leaning or multi-parent support.

To address these limitations, we present $\phi$-Scene, a physically grounded approach for open-vocabulary and compositional image-to-3D scene reconstruction.\footnote{
We focus on 3D scenes with a moderate number of rigid objects and meaningful support/contact relations.
Non-rigid/articulated objects and large scenes with a large number of objects are beyond our scope.}
Our key idea is to reconstruct the scene following its support structure: since an object rests stably only after the objects beneath it are settled, $\phi$-Scene places objects sequentially in an inferred support order, instead of optimizing all poses jointly or lifting objects top-down.

Given an image, $\phi$-Scene first uses a 3D foundation model to recover an initial 3D scene of individual objects.
It then assembles them in an inferred supporting topological order, a process we call \emph{topology-driven physical assembly}.
For each object, SDF-based optimization resolves penetrations against the pre-settled support context, and rigid-body simulation settles it under gravity and contact forces.
Because each object settles against supports that are already stable, the resulting scene is reference-aligned and in dynamic rigid-body equilibrium.
With support represented as a graph rather than a single-parent top-down tree, $\phi$-Scene also handles non-top-down relations such as leaning and multi-parent support.

We validate $\phi$-Scene across standard reconstruction metrics, perceptual studies, and dedicated physical plausibility metrics.
On 3D-Front~\cite{fu20213d} (Tab.~\ref{tab:3dfront}), $\phi$-Scene achieves the strongest overall performance among out-of-domain methods, indicating that physical grounding is an effective structural prior for recovering object transformations.
It is also consistently preferred in human/MLLM/CLIP studies (Fig.~\ref{fig:teaser},~\ref{fig:QC}; Tab.~\ref{tab:human_vlm}).
$\phi$-Scene also substantially reduces penetration and post-simulation drift as demonstrated by our dedicated physical metrics (Tab.~\ref{tab:phys_metric},~\ref{tab:pybullet}).

To our knowledge, $\phi$-Scene is among the first image-to-3D scene reconstruction methods that explicitly reaches dynamic rigid-body equilibrium while preserving reference alignment.
Because every object already rests at a stable equilibrium with physically valid contacts, the reconstructed scene is directly simulation-ready: when released in a physics engine it stays essentially in place rather than drifting or collapsing, making $\phi$-Scene readily usable in downstream applications such as simulation, robotics, and 3D gaming.

In summary, our contributions are:
\begin{itemize}
    \item We introduce $\phi$-Scene, which recasts open-vocabulary, compositional image-to-3D scene reconstruction as topology-driven physical assembly: an MLLM infers a supporting topology, and each object is settled in topological order against its already-stabilized support context via SDF-based penetration resolution followed by rigid-body simulation.
    \item We evaluate $\phi$-Scene with standard reconstruction metrics, human/MLLM/CLIP perceptual studies, and dedicated physical plausibility metrics. $\phi$-Scene ranks first among out-of-domain methods on 3D-Front, is consistently preferred by human and MLLM raters, and substantially lowers penetration and post-simulation drift, with the drift reduction confirmed on a second, independent simulator.
\end{itemize}

\section{Related Works}
\label{sec:rw}

\subsection{Image-to-3D Object}
Early holistic image-to-3D methods~\cite{qian2023magic123,wang2023imagedream,shi2023mvdream,liu2023syncdreamer} largely followed text-to-3D pipelines by combining novel-view synthesis~\cite{liu2023zero,shi2023zero123++,weng2023consistent123,chen2024cascade,liang2024luciddreamer} with test-time 3D optimization~\cite{mildenhall2021nerf,park2019deepsdf,kerbl20233d,shen2021deep}.
With the availability of large-scale 3D datasets~\cite{deitke2023objaverse,deitke2023objaverseXL}, subsequent feed-forward models further improved both quality and efficiency~\cite{liu2023one,liu2024one,long2024wonder3d,hong2023lrm,tochilkin2024triposr,xu2024instantmesh}. 

More recent holistic 3D foundation models~\cite{Hunyuan3D20,Hunyuan3D21,Hunyuan3D25,li2025triposg,seed2025seed3d,xiang2025structured,xiang2025native} learn 3D generative priors directly in native 3D latent space. 
However, they typically reconstruct 3D as single, non-compositional 3D assets.

\subsection{Image-to-3D Scene}
Early image-to-3D scene reconstruction methods~\cite{chen2024comboverse,zhou2024zero,han2025reparo,gao2024diffcad,li2024discene,gao2024graphdreamer} reconstruct each object individually and then optimize the scene layout afterward.
However, they often overlook the strong spatial coupling among objects, leading to suboptimal arrangement coherence.
More recent works~\cite{ardelean2025gen3dsr,meng2025scenegen,zhao2025depr,shi2025scenemaker} improve arrangement coherence by using multi-instance attention~\cite{huang2025midi} together with partial 3D cues such as depth maps~\cite{wang2025moge1,wang2025moge2,yang2024depth1,yang2024depth2,he2024lotus,he2025lotus,li2025depth}.
Some of them~\cite{yao2025cast,chen2025sam} further improve both capability and generalizability by training on large-scale 3D datasets~\cite{deitke2023objaverse,deitke2023objaverseXL,fu20213d,fu20213dfuture}.
Nevertheless, most existing methods still formulate 3D scene reconstruction primarily as a static prediction of 3D objects and their arrangements, without explicitly emphasizing physically grounded object interactions.

\subsection{Physically Grounded Image-to-3D Scene}
In general 3D vision, substantial effort has been devoted to producing 3D representations that are both visually plausible and physically grounded, e.g., self-supporting geometry~\cite{chen2024atlas3d,cao2026sophy,ni2024phyrecon}, articulation~\cite{yang2024physcene,luo2025physpart,liu2023few}, and physically interactive dynamics~\cite{borycki2026gasp,huang2024dreamphysics,xie2024physgaussian}.
Support reasoning has also long aided scene understanding, from classical RGB-D support inference~\cite{silberman2012indoor} to recent LLM/VLM-driven 3D scene reasoning~\cite{yang2024holodeck}, while physical compatibility has been studied for single-object modeling~\cite{guo2024physically}.

However, in compositional image-to-3D scene reconstruction, physical plausibility has only recently emerged as a promising endeavor~\cite{yao2025cast,lin2025pat3d}.
CAST~\cite{yao2025cast} uses SDF-based pose optimization, yet treats physical plausibility mainly as a static geometric constraint rather than a dynamic settling process, and jointly optimizes poses of all objects without modeling chained support dependencies, where resolving the penetration of one side may cause new penetrations of other sides.
PAT3D~\cite{lin2025pat3d} introduces rigid-body simulation, but targets text-to-3D generation rather than reference-aligned image-to-3D reconstruction, and organizes objects into a single-parent support tree, lifting each supported object above its parent's bounding box along the gravity axis before simulation; this strictly top-down design cannot represent support such as leaning or joint support by multiple objects, and further degrades under occlusions.
In short, these approaches either lack dynamic physical settling or forgo reference alignment, and none infers an explicit support structure richer than a top-down tree.

Developed concurrently\footnote{All completed around May 2026.} and independently, $\phi$-Scene, REST3D~\cite{ma2026rest3d}, and SimuScene~\cite{lee2026simuscene} make dynamic rigid-body equilibrium an explicit objective of single-image scene reconstruction, and, to the best of our knowledge, are the first to do so.
REST3D~\cite{ma2026rest3d} optimizes away violations via a gravity-support scene tree and SimuScene~\cite{lee2026simuscene} places a physics engine in the reconstruction loop, whereas $\phi$-Scene settles objects to rest through topology-driven sequential assembly while preserving reference alignment.
Both also organize support around the gravity axis: REST3D's scene tree assigns each object a single support parent, and SimuScene applies gravity-axis shape corrections without inferring an explicit support structure.
Neither therefore explicitly represents non-top-down relations such as leaning or multi-parent support, whereas $\phi$-Scene's support DAG models them directly, e.g., a ladder resting on the ground while leaning against a wall.
Overall, while physical plausibility is gaining attention, a general solution for reference-aligned and physically grounded image-to-3D scene reconstruction remains largely underexplored.

\begin{figure}
    \centering
    \includegraphics[width=\linewidth]{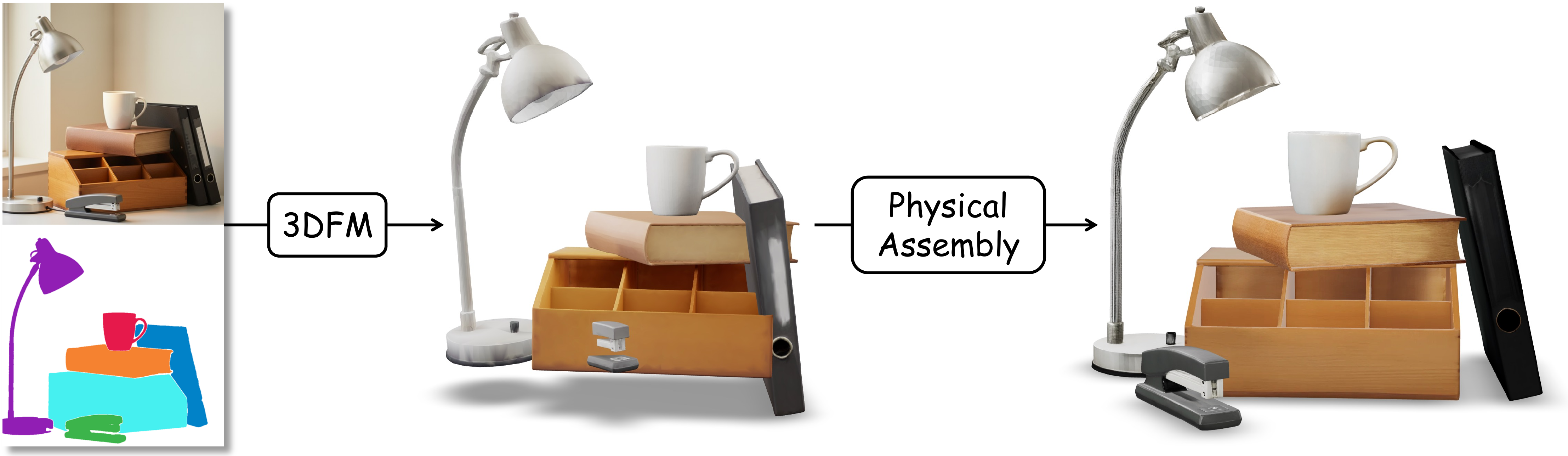}
\vspace{-.6cm}
\caption{
\textbf{Overview of $\phi$-Scene.}
From a single RGB image and its instance segmentation map, a 3D foundation model produces an initial 3D scene $\mathcal{S}^{\text{init}}$ with complete objects but a loose, physically invalid arrangement.
$\phi$-Scene then performs topology-driven physical assembly, which infers how objects support one another and sequentially settles each object against its pre-settled context, turning the initial scene into a reference-aligned and physically grounded 3D scene.
3DFM: 3D foundation model.
}
\vspace{-.3cm}
    \label{fig:stage1}
\end{figure}

\begin{figure*}
    \centering
    \includegraphics[width=\linewidth]{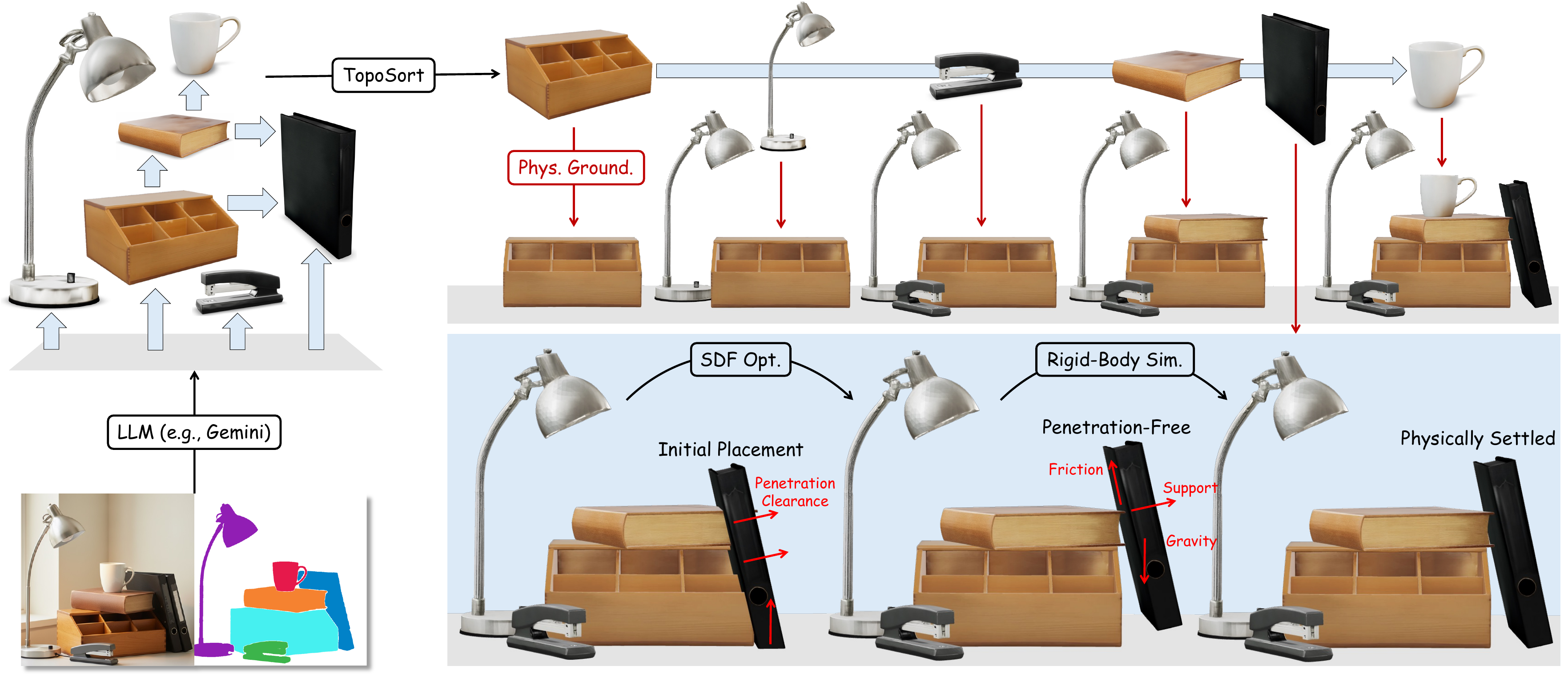}
    \vspace{-.6cm}
\caption{
\textbf{Topology-driven physical scene assembly.}
An MLLM (e.g., Gemini) infers a supporting topology over the objects and the ground plane from the image, its segmentation map, and other 3D information.
A topological sort turns it into a placement order.
Following this order, $\phi$-Scene physically grounds one object at a time against its pre-settled support context.
For each object (as shown in bottom right), SDF-based optimization first resolves penetrations into a penetration-free initial placement, and rigid-body simulation then settles it under gravity, support, and friction into a physically settled pose.
Assembling all objects in topological order yields the final physically grounded scene $\mathcal{S}^{\phi}$.
TopoSort: Topological sorting.
SDF Opt.: SDF-based penetration resolution.
}
    \vspace{-.3cm}
    \label{fig:stage23_sdf_sim}
\end{figure*}

\section{Methodology}
\label{sec:method}

Given an RGB image $I$ and its instance segmentation map $M$, $\phi$-Scene aims to reconstruct a compositional 3D scene that exhibits coherent arrangement with the reference image and physically grounded object contacts.
As shown in Fig.~\ref{fig:stage1}, $\phi$-Scene starts from an initial compositional 3D scene predicted by a 3D foundation model:
\vspace{-1pt}
\begin{equation}
\small
\mathcal{S}^{\text{init}}=
\left\{\left(\mathbf{o}_i, \mathbf{t}_i^{\text{init}}\right)\right\}_{i=1}^{N}
= \mathrm{3DFM}(I, M),
\vspace{-1pt}
\label{eq:3dfm}
\end{equation}
where $\mathbf{o}_i$ denotes the $i^{\text{th}}$ 3D object in triangular mesh and $\mathbf{t}_i^{\text{init}}$ its initial scene transformation.
$\phi$-Scene then formulates physical grounding as a topology-driven physical assembly process that infers how objects support one another and sequentially settles each object against its pre-settled support context under real-world physical constraints (Sec.~\ref{sec:stage23}).

\subsection{Topology-Driven Physical Scene Assembly}
\label{sec:stage23}
As illustrated in Fig.~\ref{fig:stage23_sdf_sim}, after initialization, $\phi$-Scene first infers a supporting topology and derives a topological order for sequential placement (Sec.~\ref{sec:vlm_sg}).
Each object is grounded against its pre-settled support context through SDF-based penetration resolution (Sec.~\ref{sec:sdf}) followed by rigid-body simulation (Sec.~\ref{sec:simulation}), ultimately yielding a physically grounded 3D scene.

\subsubsection{Supporting Topology Construction}
\label{sec:vlm_sg}
To enable support-aware sequential assembly, we first infer a supporting topology for the initial 3D scene 
\(\mathcal{S}^{\text{init}}\).
We augment the object set with a ground plane \(\mathbf{g}\), yielding 
\(\mathcal{O}^{+\mathbf{g}}=\left\{\mathbf{o}_i\right\}_{i=1}^{N}\cup\{\mathbf{g}\}\).
For each object \(\mathbf{o}_i\), we build lightweight context \(\mathbf{m}_i\) that summarizes its semantic identity \(d_i\) and its 3D bounding boxes \(\text{AABB}_i, \text{OBB}_i\) under the initial scene transformation \(\mathbf{t}_i^{\text{init}}(\mathbf{o}_i)\), with each entry linked to its region in \(M\) by the instance segmentation color.
Together with the ground node, these form the support reasoning context \(\mathcal{M}^{+\mathbf{g}}=\{\mathbf{m}_i\}_{i=1}^{N}\cup\{\mathbf{m}_{\mathbf{g}}\}\).

We infer support dependencies in a multimodal manner by feeding the RGB image \(I\), the segmentation map \(M\), and the support reasoning context \(\mathcal{M}^{+\mathbf{g}}\) into a large MLLM:
\vspace{-1pt}
\begin{equation}
\small
\label{eq:vlm_supp_graph}
f_{\text{mllm}}:(I,M,\mathcal{M}^{+\mathbf{g}})\longrightarrow \mathcal{E}^{+\mathbf{g}},
\vspace{-1pt}
\end{equation}
where \(\mathcal{E}^{+\mathbf{g}}\) is a set of directed support edges.
Each edge \((\mathbf{o}_i,\mathbf{o}_j)\in\mathcal{E}^{+\mathbf{g}}\) encodes the support relation \(\mathbf{o}_i\rightarrow\mathbf{o}_j\), meaning that \(\mathbf{o}_i\) physically supports \(\mathbf{o}_j\).
An object may be jointly supported by multiple objects, and may itself support multiple other objects.
Specific prompts and the full metadata schema are provided in Supp.~Sec.~\ref{supp:sec:suppgraph}.
Based on $\mathcal{E}^{+\mathbf{g}}$, we form the supporting topology graph $\mathcal{G}^{+\mathbf{g}}=(\mathcal{O}^{+\mathbf{g}},\mathcal{E}^{+\mathbf{g}})$.

To derive the sequential physical grounding order, we remove the ground node and obtain the object-only graph \(\mathcal{G}=(\mathcal{O},\mathcal{E})\).
We then perform Kahn's algorithm~\cite{kahn1962topological} on \(\mathcal{G}\), using descending\footnote{Larger top-down footprints often indicate base/supporting objects with more potential children. We therefore settle them earlier to provide a stable support context and avoid having their placements biased by smaller objects, which may introduce instabilities.} top-down footprint area, i.e., the area of each object's OBB projected onto the ground plane, to break ties when multiple nodes have the same zero in-degree:
\vspace{-1pt}
\begin{equation}
\small
(\mathbf{o}_{(1)},\mathbf{o}_{(2)},\dots,\mathbf{o}_{(N)})
=
\operatorname{TopoSort}\!\left(\mathcal{G}\right).
\vspace{-1pt}
\end{equation}
This order ensures that if \(\mathbf{o}_i\rightarrow \mathbf{o}_j\) and \(\mathbf{o}_i\neq \mathbf{g}\), then the supporting object \(\mathbf{o}_i\) is placed before the supported object \(\mathbf{o}_j\).
For each object \(\mathbf{o}_{(i)}\), its pre-settled support context is \(\mathcal{C}^{+\mathbf{g}}_{(i)}=\{\mathbf{g}\}\cup\left\{\mathbf{o}_{(j)}\mid j<i\right\}\).

\subsubsection{SDF-based Penetration Resolution}
\label{sec:sdf}
Before rigid-body settling, the active object \(\mathbf{o}_{(i)}\) may still penetrate its pre-settled support context \(\mathcal{C}^{+\mathbf{g}}_{(i)}\), which destabilizes contact handling in the subsequent simulation and is difficult to resolve robustly with heuristic adjustments~\cite{lin2025pat3d}.
We therefore compute a collision-free initialization by optimizing only a translation \(\mathbf{x}\in\mathbb{R}^3\) of the object (the translation component of its transformation \(\mathbf{t}\), with rotation held fixed), resolving penetrations with minimal disturbance to the arrangement coherence of \(\mathcal{S}^{\text{init}}\).
We uniformly sample surface points \(\{\mathbf{p}_n\}_{n=1}^{N_a}\) on the active mesh \(\mathbf{o}_{(i)}\) and \(\{\mathbf{q}_{k,n}\}_{n=1}^{N_k}\) on each fixed mesh \(\mathbf{o}_k\in\mathcal{C}^{+\mathbf{g}}_{(i)}\), and let \(\mathrm{SDF}(\mathbf{p},\mathbf{o})\) denote the signed point-to-mesh distance (negative inside).
We then minimize a bidirectional clearance loss that hinges the signed clearance in both directions:
\vspace{-1pt}
\begin{equation}
\small
\begin{aligned}
\mathcal{L}_{\mathrm{SDF}}(\mathbf{x})=\sum_{\mathbf{o}_k\in\mathcal{C}^{+\mathbf{g}}_{(i)}}\Big(&\sum_{n=1}^{N_a}\left[\epsilon-\mathrm{SDF}(\mathbf{p}_n+\mathbf{x},\mathbf{o}_k)\right]_++\\
&\sum_{n=1}^{N_k}\left[\epsilon-\mathrm{SDF}(\mathbf{q}_{k,n}-\mathbf{x},\mathbf{o}_{(i)})\right]_+\Big),
\end{aligned}
\vspace{-1pt}
\end{equation}
where \([\cdot]_+\!:=\!\max(\cdot,0)\), \(\epsilon\!>\!0\) is a safety clearance that penalizes both penetration and excessively small separations, and fixed-mesh points are shifted by \(-\mathbf{x}\) rather than rebuilding the translated active mesh.
The bidirectional form avoids missing penetrations on thin, sheet-like geometry.
We then optimize \(\mathbf{x}^{*}=\operatorname{argmin}_{\mathbf{x}}\mathcal{L}_{\mathrm{SDF}}(\mathbf{x})\) and translate the active-object vertices by \(\mathbf{v}'_{(i)}=\mathbf{v}_{(i)}+\mathbf{x}^{*}\), where \(\mathbf{v}_{(i)}\) denotes the vertices of \(\mathbf{o}_{(i)}\), as the collision-free initialization for rigid-body settling.

\subsubsection{Rigid-body Physical Settling}
\label{sec:simulation}
Although SDF-based penetration resolution provides a collision-free initialization for the current object \(\mathbf{o}_{(i)}\), it does not guarantee dynamic stability under gravity and contact forces.
We therefore perform rigid-body settling simulation for \(\mathbf{o}_{(i)}\) against its pre-settled support context \(\mathcal{C}^{+\mathbf{g}}_{(i)}\), allowing the object to establish stable contacts and settle into a physically grounded equilibrium.

Starting from the collision-free initialization \(\mathbf{v}'_{(i)}\) of Sec.~\ref{sec:sdf}, we simulate \(\mathbf{o}_{(i)}\) as an active rigid body while keeping the objects in \(\mathcal{C}^{+\mathbf{g}}_{(i)}\) fixed as passive support bodies.
For stability, we use simulation-oriented collision proxies rather than raw meshes (which are often non-convex): a box collider for the ground and a convex-decomposition compound for each object.
Under gravity, contact, friction, damping, and non-penetration, physically invalid placements are resolved through settling until stable contacts emerge, producing a settled transformation \(\mathbf{t}_{(i)}^{\phi}\).

To determine whether the object has settled, we further adopt a sliding-window equilibrium criterion.
Let the simulated transformation at frame \(n\) be \(\mathbf{t}_{(i)}^{n}=(\mathbf{x}_{(i)}^{n},\mathbf{r}_{(i)}^{n})\), where \(\mathbf{x}_{(i)}^{n}\) and \(\mathbf{r}_{(i)}^{n}\) denote its translation and rotation.
We declare the object physically settled when both its translational and rotational changes over a window of length \(w\) are sufficiently small:
\vspace{-1pt}
\begin{equation}
\small
\left\|\mathbf{x}_{(i)}^{n}-\mathbf{x}_{(i)}^{n-w}\right\|_2<\epsilon_{\text{pos}},\
\angle\!\left(\mathbf{r}_{(i)}^{n},\mathbf{r}_{(i)}^{n-w}\right)<\epsilon_{\text{rot}},
\vspace{-1pt}
\end{equation}
where \(\epsilon_{\text{pos}}\) and \(\epsilon_{\text{rot}}\) are the translation and rotation thresholds for declaring equilibrium.
After \(\mathbf{o}_{(i)}\) is settled, we fix its transformation as \(\mathbf{t}_{(i)}^{\phi}\) and add it to the support context for subsequent objects.

Applying SDF-based penetration resolution (Sec.~\ref{sec:sdf}) and rigid-body settling (Sec.~\ref{sec:simulation}) to all objects in the supporting topological order yields the final physically grounded scene \(\mathcal{S}^{\phi}=\left\{\left(\mathbf{o}_i,\mathbf{t}_i^{\phi}\right)\right\}_{i=1}^{N}\).

\begin{table}[t]
\centering
\scriptsize
\setlength{\heavyrulewidth}{0.08em}
\setlength{\lightrulewidth}{0.04em}
\setlength{\arrayrulewidth}{0.04em}
\setlength{\tabcolsep}{3.8pt}

\caption{
\textbf{Quantitative comparison on 3D-Front.}
$\phi$-Scene achieves the best overall performance among all out-of-domain methods and remains highly competitive with in-domain methods.
$^\circ$: Trained on 3D-Front~\cite{fu20213d}.
$^\bullet$: Trained on both 3D-Front and 3D-Future~\cite{fu20213dfuture}.
OOD: Out-of-domain.
S.: Scene.
O.: Object.
CD: Chamfer distance.
F.: F-score.
A.R.: Averaged rank.
}
\vspace{-.2cm}
\begin{tabular}{l|c|cc|cc|c|c}
\toprule
Method 
& OOD
& S. CD $\downarrow$ 
& S. F. $\uparrow$ 
& O. CD $\downarrow$ 
& O. F. $\uparrow$ 
& IoU$_{\text{bbox}}$ $\uparrow$
& A.R. $\downarrow$ \\
\midrule

SceneGen
& \ding{51}
& 0.1531 
& 0.4082 
& 0.1279 
& 0.4235 
& 0.1281
& 5.2 \\

Gen3DSR
& \ding{51}
& \underline{0.1107}
& 0.5052
& 0.1792
& 0.3780
& 0.1914
& 4.4 \\

DepR$^\bullet$
& \ding{55}
& 0.1218
& 0.5155
& 0.1607
& 0.4008
& 0.2094
& 4.2 \\

SAM3D 
& \ding{51}
& 0.1208 
& 0.4809 
& 0.1265
& 0.4598 
& 0.1625
& 4.0 \\

MIDI$^\circ$
& \ding{55}
& 0.1129
& \underline{0.5413} 
& \textbf{0.0894} 
& \textbf{0.5247} 
& \textbf{0.2532}
& \textbf{1.6} \\

$\phi$-Scene
& \ding{51}
& \textbf{0.1097} 
& \textbf{0.5438} 
& \underline{0.1081}
& \underline{0.4856}
& \underline{0.2263}
& \textbf{1.6} \\
\bottomrule
\end{tabular}

\label{tab:3dfront}
\end{table}

\begin{table}[t]
\centering
\scriptsize
\setlength{\heavyrulewidth}{0.08em}
\setlength{\lightrulewidth}{0.04em}
\setlength{\arrayrulewidth}{0.04em}
\setlength{\tabcolsep}{2.7pt}

\caption{
\textbf{Quantitative evaluation of physical plausibility.}
We evaluate physical plausibility on the same test suite used for MLLM evaluation along two complementary aspects: static contact quality, measured by penetration depth and floating distance$^{\text{\ref{fn:floating}}}$, and dynamic stability, measured by post-simulation translational and rotational drift.
$\phi$-Scene achieves the best overall performance, indicating more valid object contacts and more physically stable configurations.
P.S.: Post-simulation.
}
\vspace{-.2cm}
\begin{tabular}{l|cc|cc|c}
\toprule
Method
& Pen. Depth $\downarrow$
& Float. Dist. $\downarrow$
& P.S. Drift$_{\text{trans}}$ $\downarrow$
& P.S. Drift$_{\text{rot}}$ $\downarrow$
& A.R. $\downarrow$ \\
\midrule
MIDI
& 0.1453
& 0.0046
& 5.0545
& 89.5181$^\circ$
& 5.0 \\
SceneGen
& 0.2263
& \textbf{0.0000}
& 13.5759
& 114.1704$^\circ$
& 4.8 \\
DepR
& 0.0279
& 0.0182
& 0.3305
& \underline{53.0292}$^\circ$
& 3.5 \\
SAM3D
& 0.0305
& \underline{0.0003}
& 0.4479
& 86.7801$^\circ$
& 3.5 \\
Gen3DSR
& \underline{0.0239}
& 0.0005
& \underline{0.2949}
& 82.9112$^\circ$
& \underline{2.5} \\
$\phi$-Scene
& \textbf{0.0011}
& 0.0013
& \textbf{0.0987}
& \textbf{5.0635}$^\circ$
& \textbf{1.8} \\
\bottomrule
\end{tabular}

\label{tab:phys_metric}
\vspace{-.2cm}
\end{table}

\footnotetext{Note that some objects may not be intended to contact other objects in a 3D scene.
Therefore, floating distance should be interpreted together with penetration depth and post-simulation drift when assessing overall physical plausibility.\label{fn:floating}}

\begin{table}[t]
\centering
\scriptsize
\setlength{\heavyrulewidth}{0.08em}
\setlength{\lightrulewidth}{0.04em}
\setlength{\arrayrulewidth}{0.04em}
\setlength{\tabcolsep}{13.5pt}

\caption{
\textbf{Cross-simulator validation (PyBullet).}
We re-run the post-simulation stability protocol of Tab.~\ref{tab:phys_metric} in a separate rigid-body simulator (PyBullet) that is not used during settling.
$\phi$-Scene again attains the lowest translational and rotational drift by a large margin and the best average rank, indicating that its low drift reflects the reconstructed scene rather than the specific simulator used for settling.
Protocol details in Supp.~Sec.~\ref{supp:sec:phys_metrics}.
}

\vspace{-.2cm}
\begin{tabular}{l|cc|c}
\toprule
Method
& P.S. Drift$_{\text{trans}}$ $\downarrow$
& P.S. Drift$_{\text{rot}}$ $\downarrow$
& A.R. $\downarrow$ \\
\midrule
SceneGen
& 5.5426
& 113.2394$^\circ$
& 6.0 \\
MIDI
& 3.6252
& 81.0499$^\circ$
& 5.0 \\
Gen3DSR
& 0.6573
& 66.1770$^\circ$
& 3.0 \\
SAM3D
& 0.7344
& \underline{52.4032}$^\circ$
& 3.0 \\
DepR
& \underline{0.5992}
& 70.0730$^\circ$
& 3.0 \\
$\phi$-Scene
& \textbf{0.0405}
& \textbf{6.9597}$^\circ$
& \textbf{1.0} \\
\bottomrule
\end{tabular}

\label{tab:pybullet}
\vspace{-.2cm}
\end{table}

\newcommand{\human}{\faUser}
\newcommand{\gpt}{\faRobot}

\begin{table*}[t]
\centering
\scriptsize
\setlength{\heavyrulewidth}{0.08em}
\setlength{\lightrulewidth}{0.04em}
\setlength{\arrayrulewidth}{0.04em}
\setlength{\tabcolsep}{14pt}

\caption{
\textbf{Human, MLLM, and CLIP evaluation.}
We report CLIP$_{\text{I}}$ for image-level visual similarity between reference image and the rendered images of reconstructed 3D scenes, together with pairwise win rates (\%) evaluated by both humans and a large MLLM on visual quality, reference alignment, and physical plausibility.
$\phi$-Scene obtains the highest CLIP$_{\text{I}}$ score and the highest win rates across all human and MLLM evaluation dimensions.
}
\vspace{-.2cm}
\begin{tabular}{l|c|cc|cc|cc|c}
\toprule
Method
& CLIP$_{\text{I}}$ $\uparrow$
& VQ (\human) $\uparrow$
& VQ (\gpt) $\uparrow$
& RA (\human) $\uparrow$
& RA (\gpt) $\uparrow$
& PP (\human) $\uparrow$
& PP (\gpt) $\uparrow$
& A.R. $\downarrow$ \\
\midrule

DepR
& 0.5896
& 20.36\% & 9.47\%
& 9.46\% & 5.03\%
& 12.68\% & 9.05\%
& 5.6 \\

SceneGen
& 0.6381
& 7.50\% & 17.99\%
& 12.14\% & 22.50\%
& 7.32\% & 22.84\%
& 5.3 \\

Gen3DSR
& 0.6410
& 20.18\% & 31.94\%
& 30.00\% & 37.69\%
& 23.39\% & 33.33\%
& 4.1 \\

MIDI
& \underline{0.7693}
& 59.64\% & 75.26\%
& 55.36\% & \underline{75.38\%}
& \underline{48.57\%} & 71.73\%
& 2.6 \\

SAM3D
& 0.7356
& \underline{65.89\%} & \underline{78.49}\%
& \underline{59.29\%} & 68.34\%
& 40.36\% & \underline{73.44\%}
& \underline{2.4} \\

$\phi$-Scene
& \textbf{0.7970}
& \textbf{92.86\%} & \textbf{86.98\%}
& \textbf{93.57\%} & \textbf{91.41\%}
& \textbf{95.36\%} & \textbf{92.67\%}
& \textbf{1.0} \\

\bottomrule
\end{tabular}

\label{tab:human_vlm}
\end{table*}

\begin{table*}[t]
\centering
\scriptsize
\setlength{\heavyrulewidth}{0.08em}
\setlength{\lightrulewidth}{0.04em}
\setlength{\arrayrulewidth}{0.04em}
\setlength{\tabcolsep}{12pt}

\caption{
\textbf{Quantitative ablation.}
Only MLLM-inferred topology with sequential assembly (Ours) stays reference-aligned (highest CLIP$_{\text{I}}$) while remaining near the lowest drift, yielding the best overall rank.
J.: Joint.
S.: Sequential.
Top-down Lift.: Top-down bounding-box lifting following PAT3D~\cite{lin2025pat3d}.
Z Topo.: Heuristic topological ordering by mean vertex height.
MLLM Topo.: MLLM-inferred topology.
}
\vspace{-.2cm}
\begin{tabular}{l|c|cc|cc|c}
\toprule
Variant
& CLIP$_{\text{I}}$ $\uparrow$
& Pen. Depth $\downarrow$
& Float. Dist. $\downarrow$
& P.S. Drift$_{\text{trans}}$ $\downarrow$
& P.S. Drift$_{\text{rot}}$ $\downarrow$
& A.R. $\downarrow$ \\
\midrule
Initial Scene
& 0.7356 & 0.0305 & \textbf{0.0003} & 0.4479 & 86.7801 & 5.2 \\
J. SDF Opt. (CAST)
& 0.7601 & 0.0014 & 0.0060 & 0.2170 & 37.6154 & 4.4 \\
J. SDF Opt. + J. Sim.
& 0.7087 & 0.0027 & 0.0025 & 0.1116 & 12.4088 & 4.4 \\
Top-down Lift. + J. Sim. (PAT3D)
& 0.6584 & 0.0019 & 0.0085 & \textbf{0.0974} & \underline{4.7613} & 4.4 \\
Z Topo. + S. SDF Opt. + S. Sim.
& 0.6890 & 0.0015 & 0.0074 & 0.1015 & \textbf{4.2549} & 4.0 \\
MLLM Topo. + S. SDF Opt.
& \underline{0.7794} & \textbf{0.0007} & 0.0062 & 0.1920 & 37.2639 & \underline{3.6} \\
MLLM Topo. + S. SDF Opt. + S. Sim. (Ours)
& \textbf{0.7970} & \underline{0.0011} & \underline{0.0013} & \underline{0.0987} & 5.0635 & \textbf{2.0} \\
\bottomrule
\end{tabular}%

\label{tab:ablation}
\vspace{-.2cm}
\end{table*}

\begin{table}[t]
\centering
\scriptsize
\setlength{\heavyrulewidth}{0.08em}
\setlength{\lightrulewidth}{0.04em}
\setlength{\arrayrulewidth}{0.04em}
\setlength{\tabcolsep}{13.5pt}

\caption{\textbf{Supporting topology accuracy} on contact-rich scenes with manually annotated support edges. MLLM-inferred topology attains near-perfect precision, recall, and F1, far surpassing the height-based heuristic (Z Topo.).}
\vspace{-.2cm}
\begin{tabular}{l|ccc}
\toprule
Supporting topology & Precision $\uparrow$ & Recall $\uparrow$ & F1 $\uparrow$ \\
\midrule
Z Topo. & 0.82 & 0.69 & 0.75 \\
MLLM (Ours) & \textbf{0.98} & \textbf{0.96} & \textbf{0.97} \\
\bottomrule
\end{tabular}

\label{tab:topo}
\vspace{-.2cm}
\end{table}

\section{Experiments}
\label{sec:exp}

\subsection{Experimental Setting}
\subsubsection{Implementation Details}
We instantiate the 3D foundation model in Eq.~\ref{eq:3dfm} with SAM3D~\cite{chen2025sam}.
The supporting topology is inferred by an MLLM ({\ttfamily gemini-2.5-flash}~\cite{google_gemini_2026}).
The detailed prompting is provided in Supp.~Sec.~\ref{supp:sec:suppgraph}.
For SDF-based penetration resolution, we optimize \(\mathcal{L}_{\text{SDF}}\) using Adam~\cite{kingma2014adam} with \(N_a=N_k=2048\), \(\epsilon=0.005\), learning rate \(0.005\), and at most \(300\) iterations.
For rigid-body settling, we use Blender/BlenderProc-based sequential simulation with VHACD resolution \(1.5\times10^6\), depth \(22\), and at most \(96\) vertices per convex component.
Each object is simulated for \(120\) frames and considered settled when its pose variation over a sliding window of \(15\) frames falls below \(10^{-4}\) in translation and \(10^{-4}\,\mathrm{rad}\) in rotation.

\subsubsection{Baseline Selection}
We compare $\phi$-Scene against recent and open-sourced single image-to-3D scene methods:
\begin{enumerate}[label=\arabic*.,leftmargin=1.4em,topsep=1pt,itemsep=0pt,parsep=0pt]
    \item \textbf{SAM3D}~\cite{chen2025sam} is a visual foundation model that recovers complete 3D objects together with their (coarse) scene arrangements in a feedforward manner.
    \item \textbf{MIDI}~\cite{huang2025midi} formulates single-image 3D scene generation as a multi-instance diffusion process, denoising the latents of multiple 3D objects simultaneously with weight-shared DiTs.
    \item \textbf{SceneGen}~\cite{meng2025scenegen} jointly predicts 3D objects and their layouts in a feedforward manner, using object- and scene-level features from pretrained visual encoders.
    \item \textbf{DepR}~\cite{zhao2025depr} is a depth-guided framework that reconstructs individual objects and then optimizes their scene layouts under depth guidance.
    \item \textbf{Gen3DSR}~\cite{ardelean2025gen3dsr} is a modular framework that first parses the scene holistically using depth estimators and then reconstructs individual objects.
\end{enumerate}

\begin{figure*}[t]
    \centering
    \includegraphics[width=\linewidth]{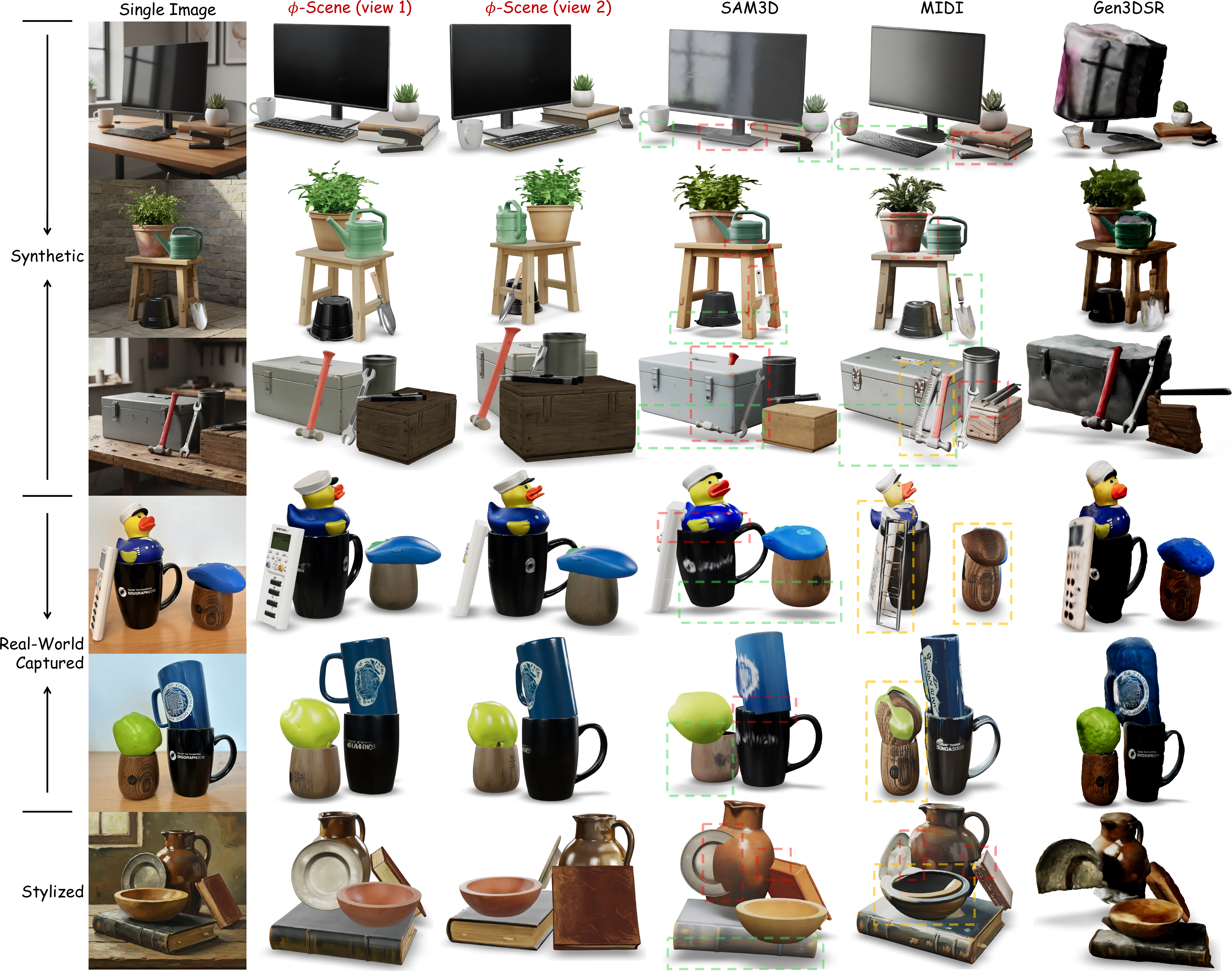}
    \vspace{-.6cm}
\caption{
\textbf{Qualitative comparison.}
Across diverse images, $\phi$-Scene produces more coherent scene arrangements and more physically valid object contacts than prior methods, with fewer penetration, floating, and unstable-contact artifacts, while preserving complete object geometry with sharper, more realistic textures.
Penetrations, floatings, and irregular objects are boxed in red, green, and yellow, respectively.
}
    \label{fig:QC}
    \vspace{-.3cm}
\end{figure*}

\subsection{Quantitative Comparison}
\subsubsection{3D-Front Benchmark}
\label{sec:3df}
Following~\cite{zhou2024zero,nie2020total3dunderstanding,huang2025midi,yao2025cast,zhao2025depr,ardelean2025gen3dsr}, we conduct quantitative evaluation on 3D-Front~\cite{fu20213d} using three groups of metrics: scene-level Chamfer Distance and F-Score, object-level Chamfer Distance and F-Score, and bounding-box IoU for scene arrangement accuracy.
We strictly follow the evaluation protocol\footnote{\url{https://github.com/VAST-AI-Research/MIDI-3D\#evaluation}} of MIDI~\cite{huang2025midi} and also test on the last 1000 samples of 3D-Front.

Note that we remove tiny objects whose masks occupy less than \(0.1\%\) of the image area, as well as objects labeled as {\ttfamily Lighting}, which are often suspended in mid-air and are therefore unsuitable for gravity-aware physical grounding.
Because the evaluated 3D-Front scenes are dominated by mostly isolated objects with weak inter-object support relations, the supporting topology is simplified to contain only ``ground \(\rightarrow\) object'' edges.

As reported in Tab.~\ref{tab:3dfront}, $\phi$-Scene achieves the strongest overall performance among out-of-domain methods and remains highly competitive with in-domain methods that are explicitly trained on 3D-Front.
Although these metrics primarily measure standard 3D reconstruction quality, the gain shows that physical grounding serves as an effective structural prior for recovering object transformations, benefiting both geometric fidelity and scene arrangement accuracy.
Specific metric calculations, as well as the reason why $\phi$-Scene and SAM3D exhibit different object-level metrics, are provided in Supp.~Sec.~\ref{supp:sec:3df}.

\subsubsection{Physical Plausibility}
\label{sec:phys_eval}
To directly quantify physical grounding, we evaluate two groups of physical plausibility metrics (Tab.~\ref{tab:phys_metric}): static contact validity, via penetration depth and floating distance, and dynamic stability, via post-simulation translational and rotational drift measured after releasing the scene in a simulator.
Small post-simulation drift indicates that the scene is already close to a stable rigid-body equilibrium.

As shown in Tab.~\ref{tab:phys_metric}, $\phi$-Scene achieves the best overall average rank across these metrics.
It substantially reduces penetration artifacts and achieves much smaller post-simulation drift, indicating that topology-driven physical assembly produces more valid contacts and more dynamically stable scene configurations.
When perturbed by an external object in a physics simulator, $\phi$-Scene reconstructions respond with plausible contacts and settling rather than drifting or collapsing, confirming they are directly simulation-ready (Supp.~Sec.~\ref{supp:sec:qual}, Fig.~\ref{fig:more_sim_demo}).
Specific metric calculations are provided in Supp.~Sec.~\ref{supp:sec:phys_metrics}.

\noindent\textbf{Cross-simulator validation.}
Post-simulation drift in Tab.~\ref{tab:phys_metric} is measured with the Blender-based rigid-body simulator used by our settling stage (Sec.~\ref{sec:simulation}), so we re-run the identical protocol in a separate simulator (PyBullet) that is not used during settling.
As reported in Tab.~\ref{tab:pybullet}, $\phi$-Scene again attains the lowest translational and rotational drift by a large margin and the best average rank, indicating that its low drift reflects the reconstructed scene itself rather than the specific simulator used for settling.

\subsubsection{Human, MLLM, and CLIP Evaluation}
\label{sec:human_vlm_clip}
Following~\cite{yao2025cast,chen2025sam,lin2025pat3d}, we further evaluate perceptual quality using CLIP image similarity (CLIP$_{\text{I}}$) and human/MLLM preferences along three dimensions: Visual Quality (VQ), Reference Alignment (RA), and Physical Plausibility (PP).
As shown in Tab.~\ref{tab:human_vlm}, $\phi$-Scene obtains the highest CLIP score and the highest human/MLLM win rates across all three dimensions.

For MLLM evaluation, we prompt {\ttfamily gemini-2.5-flash}~\cite{google_gemini_2026} on 20 scenes, rendering four fixed views (front, back, left, right) of each method into a $2\times2$ image.
Comparing 6 methods yields $6\times5=30$ ordered pairwise comparisons per scene, with swapped left/right positions to reduce positional bias; each method's win rate is the fraction of non-tied comparisons it wins.
For human evaluation, 28 participants rated 6 anonymized methods on 5 of these scenes ($1$--$5$ per dimension) in interactive 3D viewers, which we convert into the same pairwise win rate.
For CLIP, we average {\ttfamily CLIP ViT-L/14}~\cite{radford2021clip} similarities between the white-background reference image and the four rendered views.
Prompts, interfaces, and examples are in Supp.~Sec.~\ref{supp:sec:vlmeval},~\ref{supp:sec:human}.

\subsection{Qualitative Comparison}
As illustrated in Fig.~\ref{fig:teaser},~\ref{fig:QC}, across diverse images $\phi$-Scene produces more convincing 3D scenes than prior methods: it preserves complete, geometrically clean object geometry with sharper textures and more realistic materials (see Supp.~Sec.~\ref{supp:sec:texrefine} for texture refinement), and yields more coherent placements and more stable, physically plausible interactions with fewer penetration, floating, and unstable-contact artifacts.
We exclude SceneGen~\cite{meng2025scenegen} and DepR~\cite{zhao2025depr} from Fig.~\ref{fig:QC}, as their outputs are often not visually meaningful (see Supp.~Fig.~\ref{fig:QC_depr_scenegen}).

\subsection{Ablation Study}
Our design combines three choices: MLLM-inferred topology, sequential (rather than joint) optimization, and two-stage settling with SDF resolution followed by rigid-body simulation.
Tab.~\ref{tab:ablation} isolates each: all variants share the same modules and test suite and differ only in topology, order, and settling, and we read reference alignment from CLIP$_{\text{I}}$ and dynamic stability from post-simulation drift.
Most variants reach low drift, but only our full design also stays reference-aligned, giving the best rank ($2.0$ vs.\ $3.6$).

\textbf{MLLM topology.}
With sequential SDF and simulation fixed, replacing MLLM topology with the height-based heuristic (Z Topo.) keeps drift comparably low but drops CLIP$_{\text{I}}$ from $0.80$ to $0.69$.
A wrong support order does not destabilize the scene but settles it into the wrong stable configuration, as objects are grounded against misplaced supports, mirroring the topology accuracy of Sec.~\ref{sec:topo_acc}.

\textbf{Sequential vs.\ joint optimization.}
Joint assembly couples all object poses, so fixing one contact perturbs others: joint SDF (CAST~\cite{yao2025cast}), joint SDF with simulation, and top-down lifting (PAT3D~\cite{lin2025pat3d}) reach only $0.76$, $0.71$, and $0.66$ CLIP$_{\text{I}}$, far below our $0.80$.
They still drive drift down, so low drift alone is easy: what they lose is alignment, as objects move together rather than settling against a fixed, stable context.

\textbf{Two-stage physical settling.}
SDF resolution alone gives high CLIP$_{\text{I}}$ ($0.78$) and the lowest penetration but leaves large rotational drift ($37.3$), showing a collision-free placement is not yet a stable one.
Adding simulation cuts rotational drift to $5.1$ and even raises CLIP$_{\text{I}}$ to $0.80$: the two stages are complementary, with SDF removing static penetration and simulation supplying dynamic equilibrium.

\subsection{Supporting Topology Accuracy}
\label{sec:topo_acc}
On $20$ contact-rich scenes with manually annotated support edges (Tab.~\ref{tab:topo}), MLLM inference reaches $0.98/0.96/0.97$ precision/recall/F1, versus $0.82/0.69/0.75$ for the height-based heuristic (Z Topo.), with the largest gains on leaning and multi-parent support that height cannot capture.
We assume an acyclic support graph; cycles are rare ($0/20$) and resolved by dropping the lowest-confidence edge, and descending-footprint tie-breaking stabilizes larger base objects first to limit error propagation during sequential settling.

\section{Conclusion}
\label{sec:con}
We presented $\phi$-Scene, a physically grounded approach to open-vocabulary and compositional image-to-3D scene reconstruction that treats a scene not as a set of objects with predicted poses, but as a stable physical system organized by support dependencies.
Through topology-driven physical assembly, it sequentially resolves penetrations and settles objects against their pre-settled support context, keeping the scene reference-aligned while reaching a simulation-ready rigid-body equilibrium.
Experiments across reconstruction metrics, human/MLLM/CLIP evaluations, and dedicated physical plausibility metrics show that $\phi$-Scene improves both reconstruction quality and physical validity.
Its grounding still depends on the initial reconstruction quality and the collision proxies used for settling; see Supp.~Sec.~\ref{supp:sec:failurecase} for limitations and failure cases.

\newpage
\twocolumn[
\centering
\Large
\textbf{$\phi$-Scene: Physically Grounded Image-to-3D Scene Reconstruction}\\
\vspace{0.5em}Supplementary Material \\
\vspace{1.0em}
] 

\appendix

\renewcommand{\thefigure}{S\arabic{figure}}
\renewcommand{\theequation}{S\arabic{equation}}
\renewcommand{\thetable}{S\arabic{table}}

\newtcolorbox[
    auto counter,
    number freestyle={S\noexpand\arabic{\tcbcounter}}
]{promptbox}[2][]{
    title={\small Prompt~\thetcbcounter: #2},
    colback=PromptBack,
    colframe=PromptFrame,
    boxrule=0.35pt,
    left=6pt,right=6pt,top=6pt,bottom=6pt,
    breakable=true,
    fontupper=\scriptsize\ttfamily,
    #1
}
\begin{figure}[h]
    \centering
    \includegraphics[width=\linewidth]{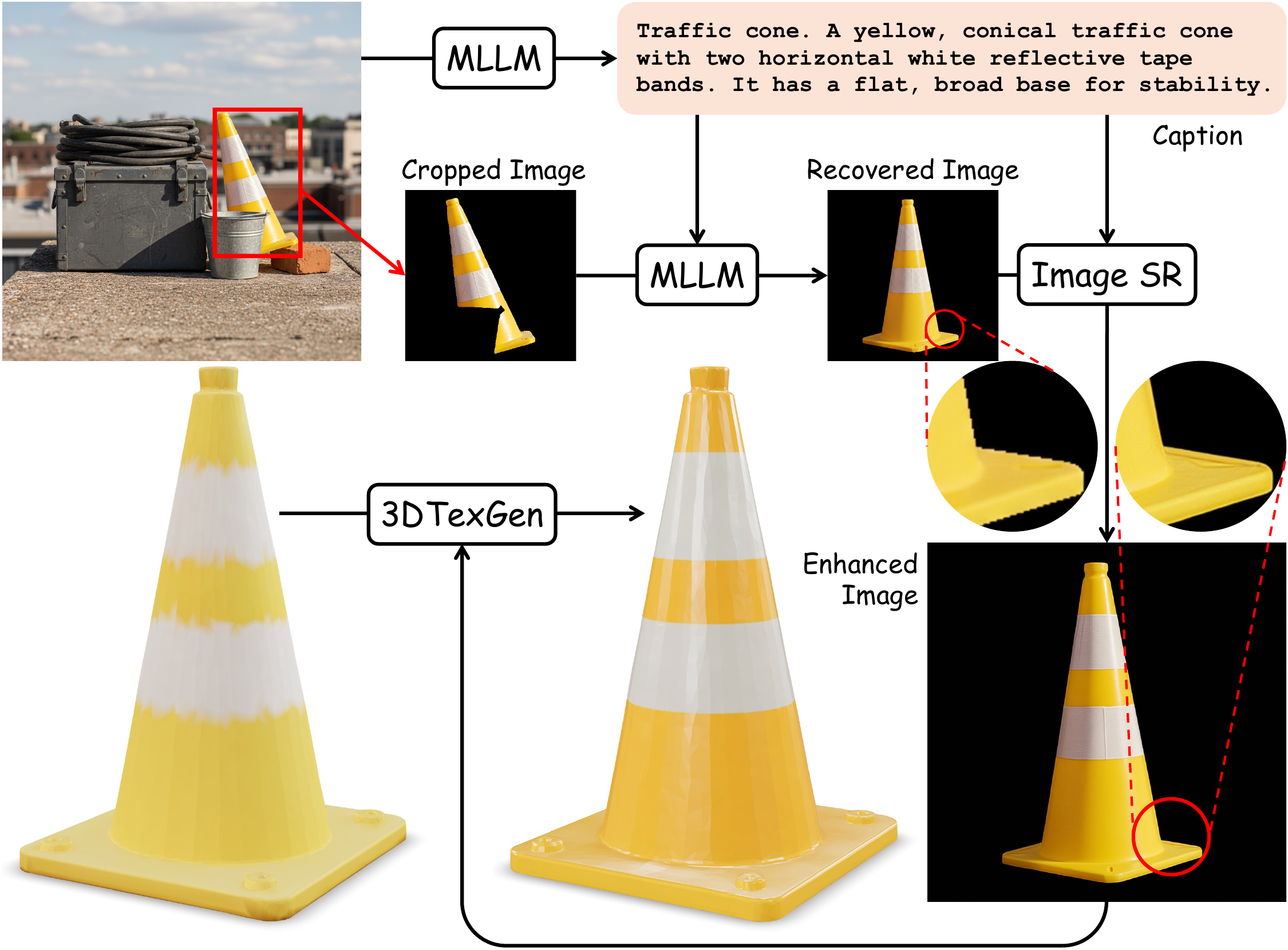}
    \vspace{-.6cm}
\caption{\textbf{Object texture refinement.}
We first highlight the object region and use Gemini~\cite{google_gemini_2026} to extract a concise object caption.
The cropped image and caption are then used to generate an isolated and complete object image, which is further enhanced by a FLUX-based~\cite{flux2024,labs2025flux1kontextflowmatching} super-resolution model. Finally, the enhanced appearance image is fed into the TRELLIS.2~\cite{xiang2025structured,xiang2025native} texturing pipeline together with the reconstructed mesh, producing sharper, more detailed, and PBR-compatible object textures.
}
    \label{supp:fig:texrefine}
\end{figure}

\begin{figure}[h]
    \centering
    \includegraphics[width=\linewidth]{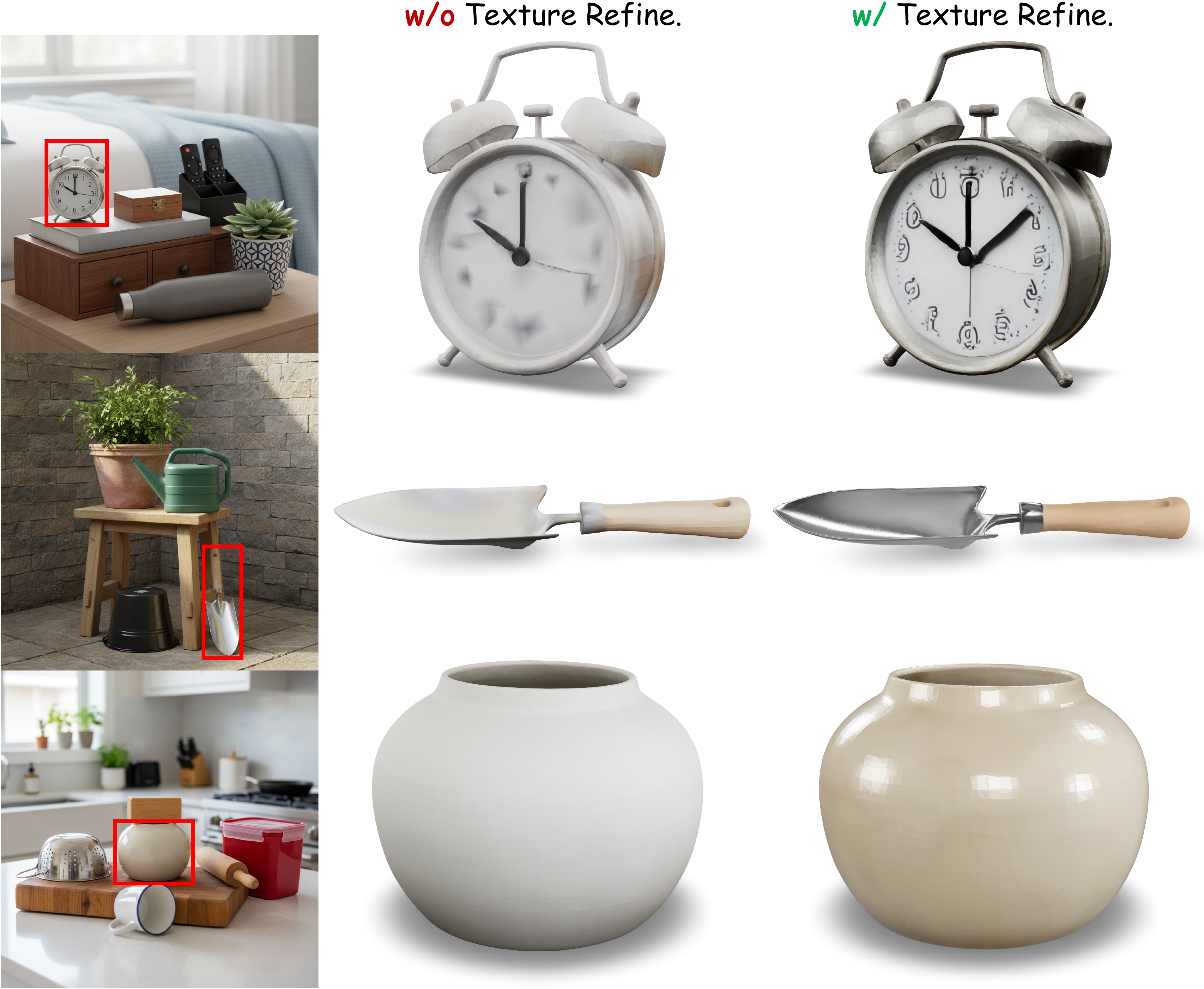}
    \vspace{-.6cm}
\caption{\textbf{Effect of object texture refinement.}
For each highlighted target object in the reference image, we compare the reconstructed 3D object before and after texture refinement.
The refinement improves visual fidelity by recovering sharper details and richer local patterns while producing PBR-compatible textures.
}
\label{supp:fig:texrefine_cases}
\end{figure}

\section{Object Texture Refinement}
\label{supp:sec:texrefine}

Beyond limited coherence with the reference image and insufficient physical plausibility, the initial 3D scene also suffers from low-quality object texture that is often overly smooth, blurry, or lacking in fine details.
Thus, $\phi$-Scene also refines the texture of each object in the initial 3D scene to obtain sharper and more detailed object textures.

Specifically, as illustrated in Fig.~\ref{supp:fig:texrefine}, for each object, we first extract a compact semantic description from the reference image using its instance mask, which serves as a semantic anchor summarizing the object's category and dominant visible appearance.
Conditioned on both this semantic description and the mask-cropped object image, we then generate a clean and complete object image that preserves the observed appearance while plausibly completing truncated or occluded regions.
The prompts for object captioning and clean object image generation are given in Prompt~\ref{supp:prompt:texrefine_1} and~\ref{supp:prompt:texrefine_2}, respectively, and both are instantiated using {\ttfamily gemini-2.5-flash}~\cite{google_gemini_2026}.
The resulting image is further enhanced by a FLUX-based~\cite{labs2025flux1kontextflowmatching,flux2024} super-resolution model\footnote{\url{https://huggingface.co/spaces/jasperai/Flux.1-dev-Controlnet-Upscaler}} to improve its spatial details and visual sharpness.
Finally, we feed the reconstructed object mesh together with the high-resolution appearance image into the texturing pipeline of TRELLIS.2~\cite{xiang2025structured,xiang2025native}, which synthesizes detailed textures with PBR-compatible material cues.

As shown in Fig.~\ref{supp:fig:texrefine_cases}, this refinement consistently improves texture quality, yielding sharper local patterns, clearer material appearance, and richer details.
In this way, $\phi$-Scene reconstructs 3D scenes that are not only coherent with the reference image and physically grounded, but also exhibit higher visual quality with rich, detailed, and PBR-compatible textures.

\begin{promptbox}[label={supp:prompt:texrefine_1}]{For object captioning.}
You are given an image: the original scene with a red bounding box highlighting a specific object.
If multiple objects are visible inside the red bounding box, describe only the primary object, ignoring surrounding objects, background context, and decorative elements.\\\\
Return format: \{Object category\}. \{Object description\}.\\\\
Be concise and precise. The description should be within 50 words.
\end{promptbox}

\begin{promptbox}[label={supp:prompt:texrefine_2}]{For clean, complete object image generation.}
Generate a clean and complete image of only the primary object in the input image.
Preserve the observed appearance of the object as much as possible.
If some parts are missing, occluded, or incomplete, plausibly complete only those missing regions.
Return only a single isolated object on an empty dark background. No ground, no surrounding objects, no decorations, and no additional elements.\\\\
Description: \dyn{Object caption extracted in the previous step}
\end{promptbox}

\section{Prompts for Supporting Topology Inference}
\label{supp:sec:suppgraph}

Following Eq.~\ref{eq:vlm_supp_graph} in Sec.~\ref{sec:vlm_sg}, we provide the detailed prompts used for MLLM-based supporting topology inference (Prompt~\ref{supp:prompt:vlm_sg_1}-\ref{supp:prompt:vlm_sg_2}). 
The underlying MLLM is {\ttfamily gemini-2.5-flash}~\cite{google_gemini_2026}.

\begin{promptbox}[label={supp:prompt:vlm_sg_1}]{System prompt for Supporting Topology Inference}
You are a careful 3D scene reasoning assistant.
You will infer object support dependencies from a scene image, a color segmentation image, and scene metadata containing object names, colors, and oriented bounding boxes.
The ground plane must be included as a valid support object.
Every object listed in the metadata must appear in your output.
Return strict JSON only.
\end{promptbox}

\begin{promptbox}[label={supp:prompt:vlm_sg_2}]{User prompt for Supporting Topology Inference}
You are given:\\
1. A scene image.\\
2. A segmentation image where each object is shown in a unique flat color.\\
3. A JSON file including each object's description, instance RGB color, and metadata (OBB, AABB, Footprint).\\\\
Task:\\
- Infer support dependencies as directed edges from the supporting object to the supported object.\\
- Include the ground plane if it supports objects.\\
- Use the segmentation colors and metadata object names exactly as given.\\
- Prefer physically plausible load-bearing support rather than merely visual overlap.\\
- A single object may support multiple objects, and a single object may also be supported by multiple objects simultaneously.
If such configurations are physically plausible, return all corresponding support edges.\\
- Support includes all physically meaningful contact relations, not only being vertically on top.
This includes any contact that contributes to an object's physical support or stability, such as leaning contact and other non-vertical contact.\\
- If object A is supported by object B, output an edge B -> A.\\
- If you cannot determine the support source for an object (e.g., it is fully occluded),
  assume it rests on the ground and add a ground -> object edge with low confidence.\\\\
Return strict JSON with this schema:\\
\{\\
\hspace*{1em}"objects": ["Plane", "object\_01", "..."],\\
\hspace*{1em}"edges": [\\
\hspace*{2em}\{\\
\hspace*{3em}"from": "supporting\_object\_name",\\
\hspace*{3em}"to": "supported\_object\_name",\\
\hspace*{3em}"relation": "supports",\\
\hspace*{3em}"confidence": 0.0--1.0,\\
\hspace*{3em}"reason": "short explanation"\\
\hspace*{2em}\}\\
\hspace*{1em}]\\
\}\\\\
Object list (every name below must appear in "objects"):\\
- object\_01\\
- object\_02\\
- ...
\dyn{All object names in the given JSON file}
\end{promptbox}

\section{3D-Front Benchmark Metrics}
\label{supp:sec:3df}

For each test sample in 3D-Front~\cite{fu20213d}, we report five metrics, as reported in Tab.~\ref{tab:3dfront}: scene-level chamfer distance and F-score, object-level chamfer distance and F-score, and bounding-box IoU.
This section briefly describes how these metrics are computed.

\subsection{Scene-Level 3D Reconstruction}

Let \(\hat{\mathcal{S}}\) and \(\mathcal{S}^{\ast}\) denote the predicted and GT 3D scenes, respectively.
Before computing the scene-level metrics, \(\hat{\mathcal{S}}\) and \(\mathcal{S}^{\ast}\) should be well-aligned.
Specifically, we first match the scene centroids and apply a uniform rescaling to the predicted scene:
\begin{equation}
\label{supp:eq:scenealign_1}
\small
\hat{\mathbf{v}}'
=
e^{\mathrm{rel}}\left(\hat{\mathbf{v}}-\hat{\mathbf{c}}\right)
+
\mathbf{c}^{\ast},
\end{equation}
where \(\hat{\mathbf{v}}\) denotes the vertices of \(\hat{\mathcal{S}}\), \(\hat{\mathbf{c}}\) and \(\mathbf{c}^{\ast}\) denote the centroids of the predicted and GT 3D scene, and \(e^{\mathrm{rel}}=
\operatorname{median}
\left(
\frac{\operatorname{sort}\left(\mathbf{e}_i^{\ast}\right)}
     {\operatorname{sort}\left(\hat{\mathbf{e}}\right)}
\right)\) is the relative scale factor.
We then further align the predicted scene to GT scene by rigid ICP~\cite{besl1992method}:
\begin{equation}
\label{supp:eq:scenealign_2}
\small
\hat{\mathcal{S}}''=\operatorname{ICP}\!\left(\hat{\mathcal{S}}'\rightarrow\mathcal{S}^{\ast}\right),
\end{equation}
where $\hat{\mathcal{S}}'$ is the predicted 3D scene after centroid matching and uniform rescaling.
For notation simplicity, we continue to use \(\hat{\mathcal{S}}\) to denote the predicted 3D scene after ICP alignment.

Let \(\{\hat{\mathbf{p}}_n\}_{n=1}^{\hat{N}}\) and \(\{\mathbf{p}_m^{\ast}\}_{m=1}^{N^{\ast}}\) denote the surface points sampled from \(\hat{\mathcal{S}}\) and \(\mathcal{S}^{\ast}\), respectively ($\hat{N}=N^{\ast}=20480$).
The bidirectional scene-level chamfer distances are defined as:
\begin{equation}
\label{supp:eq:cd}
\small
\begin{aligned}
d_{\hat{\mathcal{S}}\rightarrow\mathcal{S}^{\ast}}
&=
\frac{1}{\hat{N}}
\sum_{n=1}^{\hat{N}}
\min_{1\le m\le N^{\ast}}
\left\|
\hat{\mathbf{p}}_n-\mathbf{p}_m^{\ast}
\right\|_2^2,\\
d_{\mathcal{S}^{\ast}\rightarrow\hat{\mathcal{S}}}
&=
\frac{1}{N^{\ast}}
\sum_{m=1}^{N^{\ast}}
\min_{1\le n\le \hat{N}}
\left\|
\mathbf{p}_m^{\ast}-\hat{\mathbf{p}}_n
\right\|_2^2.
\end{aligned}
\end{equation}
Accordingly, the scene-level chamfer distance is: $d_{\hat{\mathcal{S}}\rightarrow\mathcal{S}^{\ast}}+d_{\mathcal{S}^{\ast}\rightarrow\hat{\mathcal{S}}}$.

The scene-level F-score is computed from the corresponding precision and recall under a distance threshold \(\tau=0.1\):
\begin{equation}
\label{supp:eq:fscore}
\small
\begin{aligned}
\mathrm{P}
&=
\frac{1}{\hat{N}}
\sum_{n=1}^{\hat{N}}
\left[1\ \text{if} \
\left(\min_{1\le m\le N^{\ast}}
\left\|
\hat{\mathbf{p}}_n-\mathbf{p}_m^{\ast}
\right\|_2
<\tau\right)
\ \text{else} \ 0
\right],
\\
\mathrm{R}
&=
\frac{1}{N^{\ast}}
\sum_{m=1}^{N^{\ast}}
\left[1\ \text{if} \
\left(
\min_{1\le n\le \hat{N}}
\left\|
\mathbf{p}_m^{\ast}-\hat{\mathbf{p}}_n
\right\|_2
<\tau\right)
\ \text{else} \ 0
\right].
\end{aligned}
\end{equation}
Thus, the final scene-level F-score is $\frac{2\mathrm{P}\mathrm{R}}{\mathrm{P}+\mathrm{R}+10^{-8}}$.

\subsection{Object-Level 3D Reconstruction}

Let \(\hat{\mathbf{o}}_k\) and \(\mathbf{o}_k^{\ast}\) denote the predicted and GT 3D object for the \(k\)-th object, respectively.
To isolate intrinsic object shape quality from pose and scale, we first normalize \(\hat{\mathbf{o}}_k\) and \(\mathbf{o}_k^{\ast}\) independently into the canonical cube \([-0.95,0.95]^3\), and then further align the normalized predicted object to the normalized GT object by rigid ICP~\cite{besl1992method}:
\begin{equation}
\small
\hat{\mathbf{o}}_k'=
\operatorname{ICP}\!\left(\hat{\mathbf{o}}_k\rightarrow\mathbf{o}_k^{\ast}\right).
\end{equation}
We then compute the object-level chamfer distance and F-score similar to Eq.~\ref{supp:eq:cd} and Eq.~\ref{supp:eq:fscore}, respectively, where 20480 surface points are sampled from \(\hat{\mathbf{o}}_k'\) and \(\mathbf{o}_k^{\ast}\).

Note that the object-level metrics of $\phi$-Scene are not expected to exactly match those of SAM3D~\cite{chen2025sam}, even though $\phi$-Scene uses its object geometries as initialization.
This is because the subsequent two-stage physical grounding of $\phi$-Scene further updates each object's 6-DoF pose, especially its rotation, while the evaluation protocol normalizes each object independently by per-axis min/max scaling.
Such normalization is anisotropic and rotation-sensitive, and may therefore map the same mesh under different orientations to geometrically different normalized shapes.
This discrepancy cannot be fully removed by the subsequent rigid ICP, leading to different object-level metric values.

\subsection{Scene Arrangement Coherence}

To evaluate scene-level object arrangement, we first align the predicted and GT scenes by Eq.~\ref{supp:eq:scenealign_1}-\ref{supp:eq:scenealign_2}.
We then extract the \(k\)-th predicted and GT objects from the aligned scenes, denoted by \(\hat{\mathbf{o}}_k\) and \(\mathbf{o}_k^{\ast}\), respectively, and compute their AABBs:
\begin{equation}
\small
\hat{\mathbf{b}}_k=
\left(\hat{\mathbf{b}}_{k,\min},\hat{\mathbf{b}}_{k,\max}\right),\
\mathbf{b}_k^{\ast}=
\left(\mathbf{b}_{k,\min}^{\ast},\mathbf{b}_{k,\max}^{\ast}\right).
\end{equation}

Then, we denote the intersection volume between the two bounding boxes as $V_k^{\cap}$, and the corresponding union volume as $V_k^{\cup}$.
Accordingly, bounding-box IoU is computed as $\frac{V_k^{\cap}}{V_k^{\cup}+10^{-8}}$, which reflects the accuracy of scene arrangement.
\section{Physical Plausibility Metrics}
\label{supp:sec:phys_metrics}

As reported in Tab.~\ref{tab:phys_metric} and~\ref{tab:pybullet}, we evaluate the physical plausibility of each reconstructed 3D scene from two complementary perspectives: static contact validity and dynamic stability.
Static contact validity is measured by penetration depth and floating distance, which quantify penetration and floating artifacts at object interfaces.
Dynamic stability is measured by post-simulation translational and rotational drift, which quantify how much the reconstructed scene changes after being released in a rigid-body simulator.
This section describes how these metrics are computed.

\subsection{Static Contact Validity}

For each reconstructed 3D scene \(\hat{\mathcal{S}}=\{\hat{\mathbf{o}}_i\}_{i=1}^{K}\), we first evaluate static contact validity.
A physically valid contact configuration should avoid both inter-object penetration and unrealistic floating gaps.

For each object pair \((\hat{\mathbf{o}}_i,\hat{\mathbf{o}}_j)\), \(i\neq j\), we uniformly sample \(N=2048\) surface points from both meshes, denoted by \(\{\hat{\mathbf{p}}_{i,n}\}_{n=1}^{N}\) and \(\{\hat{\mathbf{p}}_{j,m}\}_{m=1}^{N}\), respectively.
Following the signed distance definition in Sec.~\ref{sec:sdf}, we compute the bidirectional signed distances:
\begin{equation}
\small
s_{n}^{i\rightarrow j}
=
\operatorname{SDF}\!\left(\hat{\mathbf{p}}_{i,n},\hat{\mathbf{o}}_j\right),
\quad
s_{m}^{j\rightarrow i}
=
\operatorname{SDF}\!\left(\hat{\mathbf{p}}_{j,m},\hat{\mathbf{o}}_i\right).
\end{equation}
Then, the worst bidirectional signed distance for this pair is:
\begin{equation}
\small
w_{ij}
=
\min\!\left(
\min_{1\le n\le N} s_{n}^{i\rightarrow j},
\min_{1\le m\le N} s_{m}^{j\rightarrow i}
\right).
\end{equation}
Here, \(w_{ij}<0\) indicates inter-object penetration, while \(w_{ij}>0\) indicates a positive gap between the two objects.

For each object \(\hat{\mathbf{o}}_i\), we summarize its closest contact status with respect to all surrounding objects as:
\begin{equation}
\small
w_i=\min_{j\neq i} w_{ij}.
\end{equation}
The object-level penetration depth and floating distance are then defined as:
\begin{equation}
\small
d_i^{\mathrm{pen}}
=
\max(0,-w_i),
\
d_i^{\mathrm{float}}
=
\max(0,w_i).
\end{equation}
Thus, \(d_i^{\mathrm{pen}}>0\) means that \(\hat{\mathbf{o}}_i\) penetrates at least one surrounding object, while \(d_i^{\mathrm{float}}>0\) measures the positive gap from \(\hat{\mathbf{o}}_i\) to its closest surrounding object when no penetration is detected, indicating a potential floating artifact or loose contact.
Note that some objects may not be intended to contact other objects in a 3D scene.
Therefore, floating distance should be interpreted together with penetration depth and post-simulation drift when assessing overall physical plausibility.

The scene-level penetration depth and floating distance are obtained by averaging over all objects:
\begin{equation}
\small
d^{\mathrm{pen}}_{\hat{\mathcal{S}}}
=
\frac{1}{K}\sum_{i=1}^{K} d_i^{\mathrm{pen}},
\ 
d^{\mathrm{float}}_{\hat{\mathcal{S}}}
=
\frac{1}{K}\sum_{i=1}^{K} d_i^{\mathrm{float}}.
\end{equation}
The final penetration depth and floating distance of each method are obtained by averaging \(d^{\mathrm{pen}}_{\hat{\mathcal{S}}}\) and \(d^{\mathrm{float}}_{\hat{\mathcal{S}}}\) over all test scenes.

\subsection{Post-Simulation Stability}

Static contact validity alone does not guarantee that the reconstructed scene is dynamically stable.
We therefore further evaluate post-simulation stability by releasing all reconstructed objects in a rigid-body simulator and measuring how much they drift from their initial reconstructed poses.

For each reconstructed scene \(\hat{\mathcal{S}}\), all foreground objects are simulated simultaneously under the same physical rules as in Sec.~\ref{sec:simulation}, including contact, gravity, friction, damping, and non-penetration.
Before simulation, we construct collision proxies following Sec.~\ref{sec:simulation}, place a synthetic ground plane right below the scene and apply a small upward offset to all active objects so that they can settle downward into contact.

Let the pre-simulation pose of object \(\hat{\mathbf{o}}_i\) be \((\hat{\mathbf{x}}_i,\hat{\mathbf{r}}_i)\), and let its post-simulation pose be \((\tilde{\mathbf{x}}_i,\tilde{\mathbf{r}}_i)\).
The translational and rotational drift are defined as:
\begin{equation}
\small
d_i^{\mathrm{trans}}
=
\left\|
\tilde{\mathbf{x}}_i-\hat{\mathbf{x}}_i
\right\|_2,
\ 
d_i^{\mathrm{rot}}
=
\angle\!\left(\tilde{\mathbf{r}}_i,\hat{\mathbf{r}}_i\right),
\end{equation}
where \(\angle(\cdot,\cdot)\) denotes the relative rotation angle.

Since the upward offset introduces a small systematic downward displacement, we correct the translational drift by subtracting this offset, while the rotational drift is computed directly from the relative rotation.
All methods are evaluated under the same simulation setup for \(120\) frames.
The scene-level post-simulation translational and rotational drift are obtained by averaging \(d_i^{\mathrm{trans}}\) and \(d_i^{\mathrm{rot}}\) over all foreground objects, and the final method-level scores are obtained by averaging over all test scenes.
Lower post-simulation drift indicates that the reconstructed scene is closer to a stable rigid-body equilibrium.

\begin{figure*}
    \centering
    \includegraphics[width=\linewidth]{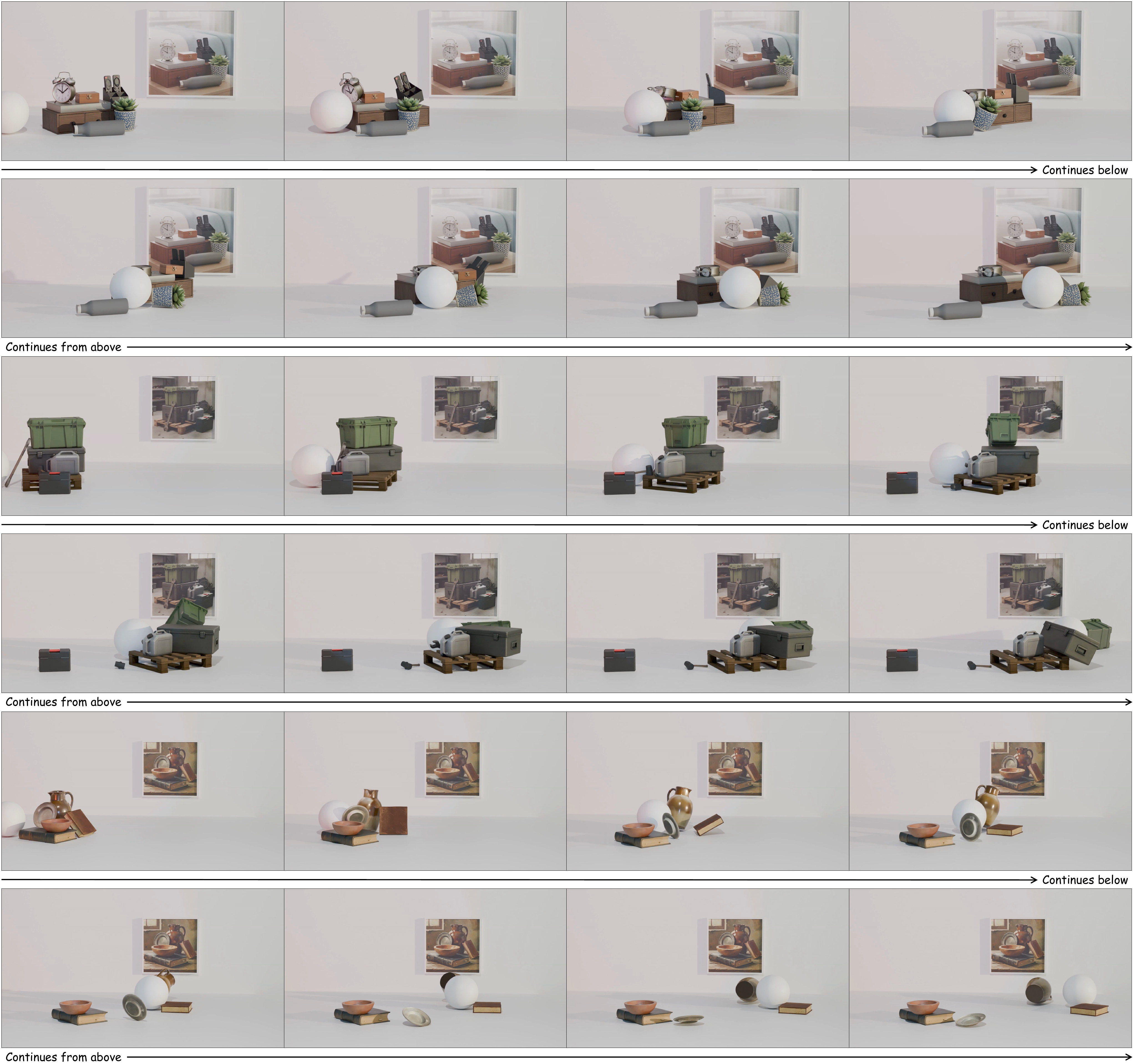}
    \vspace{-.6cm}
\caption{
\textbf{Simulation-ready 3D scene reconstruction.}
We import a scene reconstructed by $\phi$-Scene into a physics simulator and apply an external dynamic perturbation using a textured ball.
The reconstructed objects respond through plausible contacts, collisions, motion, and settling, demonstrating that $\phi$-Scene produces physically grounded 3D scenes suitable for downstream simulation.
}
    \label{fig:more_sim_demo}
\end{figure*}

\begin{figure*}
    \centering
    \includegraphics[width=\linewidth]{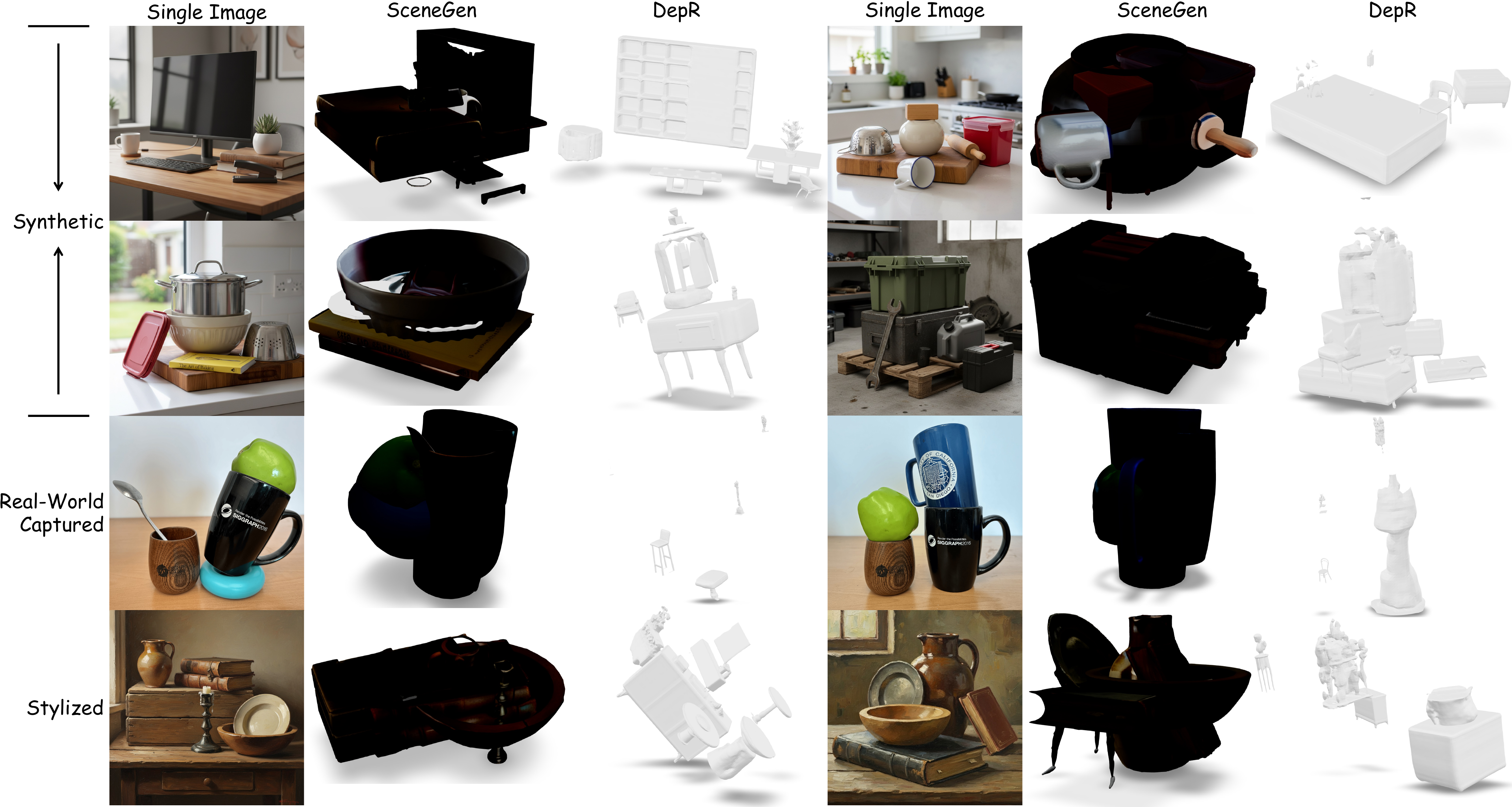}
    \vspace{-.6cm}
\caption{
\textbf{Additional qualitative results of SceneGen and DepR}.
These results are often not visually meaningful, frequently exhibiting collapsed geometry, missing structures, or unrecognizable scene compositions.
We therefore present them separately from the main qualitative comparison.
}
    \label{fig:QC_depr_scenegen}
\end{figure*}

\section{Additional Qualitative Results}
\label{supp:sec:qual}

We provide additional qualitative results that complement the main paper.
Fig.~\ref{fig:more_sim_demo} presents simulation-ready examples: each $\phi$-Scene reconstruction is imported into a physics simulator and perturbed by an external object, and the scene responds with plausible contacts, collisions, and settling rather than drifting or collapsing.

Fig.~\ref{fig:QC_depr_scenegen} instead shows representative outputs of SceneGen~\cite{meng2025scenegen} and DepR~\cite{zhao2025depr}; as these are often not visually meaningful, we exclude them from the side-by-side comparison in the main paper.

\definecolor{myred}{RGB}{180, 50, 50}
\definecolor{mygreen}{RGB}{40, 130, 90}
\definecolor{mypurple}{RGB}{130, 70, 170}

\lstdefinestyle{promptcode}{
    language=Python,
    basicstyle=\ttfamily\scriptsize,
    breaklines=true,
    breakatwhitespace=false,
    columns=fullflexible,
    keepspaces=true,
    showstringspaces=false,
    frame=none,
    numbers=none,
    linewidth=\linewidth,
    escapeinside={(*@}{@*)}
}

\section{MLLM Evaluation Prompts and Examples}
\label{supp:sec:vlmeval}

Following Sec.~\ref{sec:exp} and prompt templates from GPTEval3D~\cite{wu2024gpt}, we provide the main prompt (Prompt~\ref{supp:prompt:vlm_eval}) and one prompt for each of the three evaluation dimensions (Prompt~\ref{supp:prompt:LLM_visual}-\ref{supp:prompt:LLM_physical}) used for MLLM evaluation based on {\ttfamily gemini-2.5-flash}~\cite{google_gemini_2026}.

\begin{promptbox}[label={supp:prompt:vlm_eval}]{Main Prompt for MLLM Evaluation}
\sloppy
You will receive two images for a pairwise 3D scene evaluation.\\
Image 1 is the REFERENCE IMAGE: the real scene photo that both generated \\
3D scenes were generated from.\\\\
\dyn{Reference image}\\\\
Image 2 is the COMPARISON IMAGE:\\
- LEFT half = Scene 1 \\
- RIGHT half = Scene 2 \\
- A black vertical line separates the two scenes\\
- Each scene is shown from four views: front, back, left, and right\\
Inspect all four views of both scenes before answering.\\
Compare the overall LEFT scene against the overall RIGHT scene.\\

\dyn{Rendered scene pair image}\\

Now follow the evaluation instructions below exactly.\\
\dyn{Criterion name and its evaluation description}\\

Use the reference image and all four rendered views to evaluate the criterion. Base your judgment only on visible evidence.\\

\# Decision Options\\
Choose one of the following:\\
1. Left scene is better\\
2. Right scene is better\\
3. Cannot decide\\

Use "Cannot decide" only when both scenes perform similarly under the criterion, or when the available visual evidence is insufficient to distinguish them.\\

\# Output Format\\
Provide a concise analysis and final decision in the following format:\\

\dyn{Criterion name}:\\
Left: \{Briefly describe the left scene with respect to this criterion, including major strengths and issues.\}\\
Right: \{Briefly describe the right scene with respect to this criterion, including major strengths and issues.\}\\
Comparison: \{Explain which scene is better according to this criterion and why.\}\\

Final answer:\\
\{1, 2, or 3\}
\end{promptbox}

\begin{promptbox}[label={supp:prompt:LLM_visual}]{Prompt for Visual Quality}
\sloppy
\# Evaluation Criterion: Visual Quality\\

Visual quality evaluates how visually coherent, complete, and artifact-free the generated 3D scene appears.\\

Focus on the following observable attributes:\\
- Whether the scene has clean geometry and recognizable object shapes\\
- Whether surfaces, textures, and colors appear coherent\\
- Whether the rendering is free from severe artifacts, holes, distortions, or broken parts\\
- Whether object boundaries and structures are clear\\
- Whether the scene looks visually complete from multiple views\\
- Whether the overall appearance is consistent across the four rendered views\\

Do NOT judge the scenes based on:\\
- Reference alignment, unless the mismatch also affects visual coherence\\
- Physical plausibility, unless the issue creates visible artifacts\\
- Personal aesthetic preference beyond visible quality and coherence
\end{promptbox}

\begin{promptbox}[label={supp:prompt:LLM_reference}]{Prompt for Reference Alignment}
\sloppy
\# Evaluation Criterion: Reference Alignment\\

Reference alignment evaluates how well the generated 3D scene matches the reference image.\\

Focus on the following observable attributes:\\
- Whether the main objects in the reference image are present\\
- Whether important objects are missing, duplicated, or hallucinated\\
- Whether object shapes, silhouettes, and structures match the reference\\
- Whether the spatial layout and relative object positions match the reference\\
- Whether object scale and proportions are consistent with the reference\\
- Whether colors, textures, materials, and distinctive details are preserved\\

Do NOT judge the scenes based on:\\
- General visual quality or rendering sharpness, unless it affects reference matching\\
- Physical plausibility, unless it changes whether the scene matches the reference\\
- Personal aesthetic preference beyond reference alignment
\end{promptbox}

\begin{promptbox}[label={supp:prompt:LLM_physical}]{Prompt for Physical Plausibility}
\sloppy

\# Evaluation Criterion: Physical Plausibility\\

Physical plausibility evaluates whether the generated 3D scene appears physically possible and consistent with basic real-world constraints.\\

Focus on the following observable attributes:\\
- Whether objects adhere to the laws of physics, such as gravity, friction, and damping\\
- Whether objects interpenetrate, clip through, or overlap in impossible ways\\
- Whether object placement appears stable and physically possible\\
- Whether severe geometry artifacts make the structure physically implausible\\

Do NOT judge the scenes based on:\\
- Reference alignment, unless a mismatch causes physical impossibility\\
- General visual quality, texture quality, or aesthetics
\end{promptbox}

Motivated by the empirical noisiness of MLLM judgments, we query the API three times for each pairwise comparison under each evaluation criterion and aggregate the results by majority voting.
For each API call, we assign a vote of $+1$ if the MLLM selects the left result, $-1$ if it selects the right result, and $0$ if it outputs ``cannot decide''.
Then, we sum the votes.
A positive total vote is counted as left, a negative total vote as a right, and a zero total vote as a tie.

\begin{figure}[H]
    \centering
    \includegraphics[width=\linewidth]{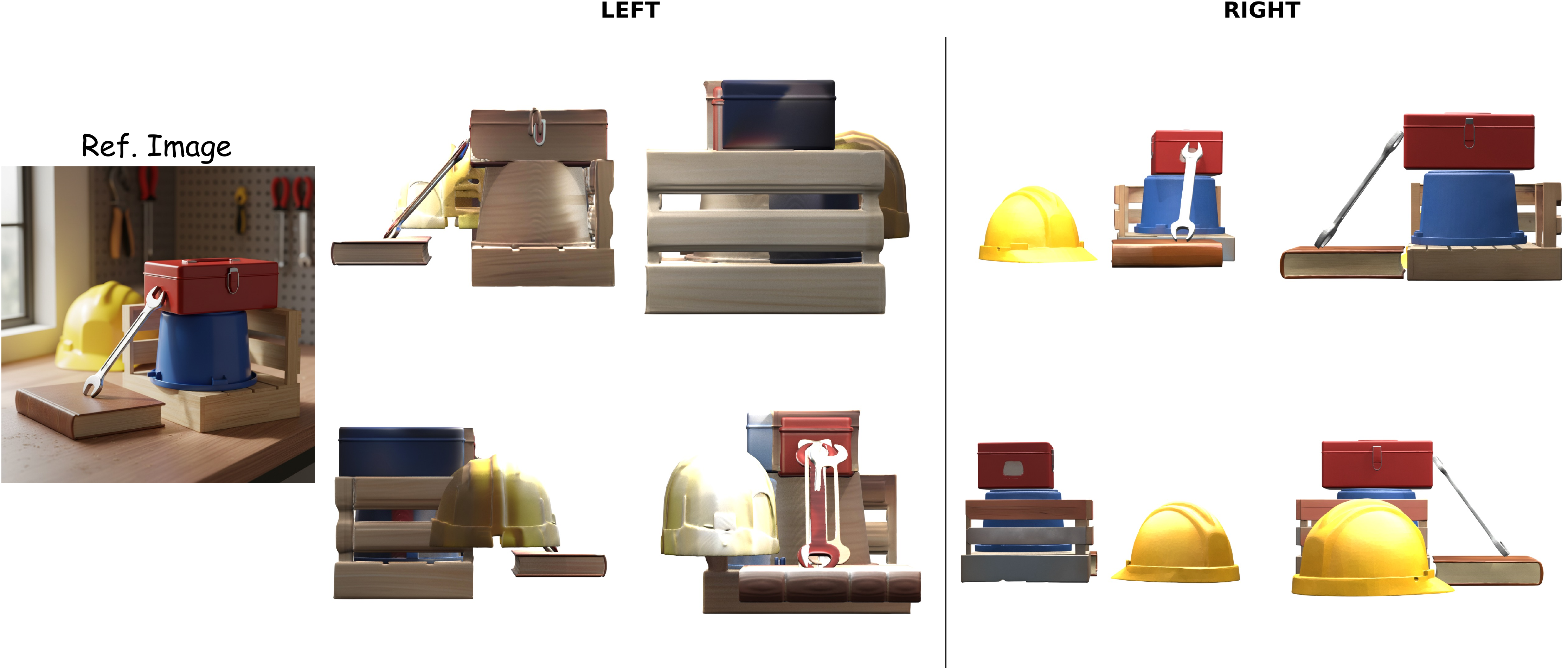}
    \vspace{-.6cm}
    \caption{\textbf{Left: MIDI. Right: $\phi$-Scene.} In visual quality evaluation, $\phi$-Scene produces cleaner geometry, more recognizable object shapes, and fewer severe rendering artifacts across views compared to MIDI.}
    \label{supp:fig:midi_vs_phi-scene}
\end{figure}

\begin{figure}[H]
    \centering
    \includegraphics[width=\linewidth]{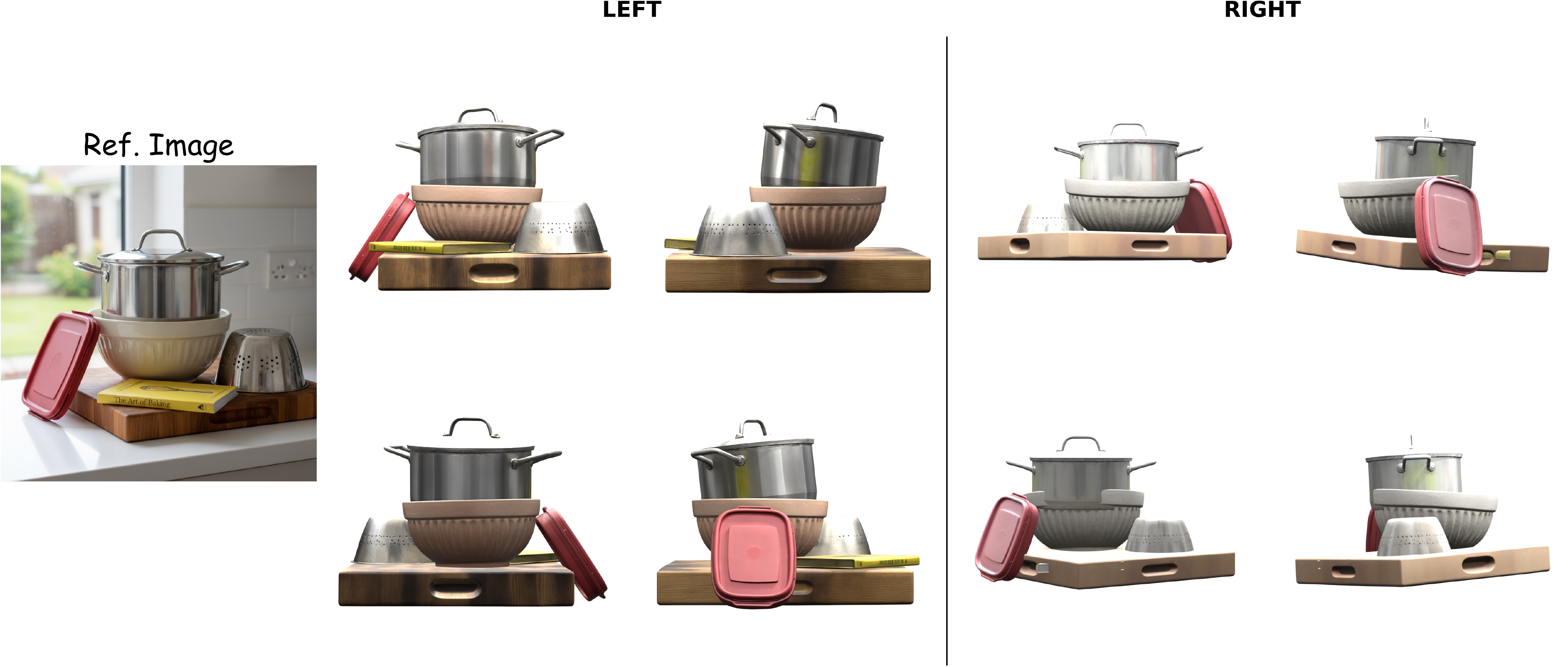}
    \vspace{-.6cm}
    \caption{\textbf{Left: $\phi$-Scene. Right: SAM3D.} In reference alignment evaluation, $\phi$-Scene more closely matches the reference image in object presence, spatial layout, and relative object placement compared to SAM3D.}
    \label{supp:fig:phi-scene_vs_sam3d}
\end{figure}

\begin{figure}[H]
    \centering
    \includegraphics[width=\linewidth]{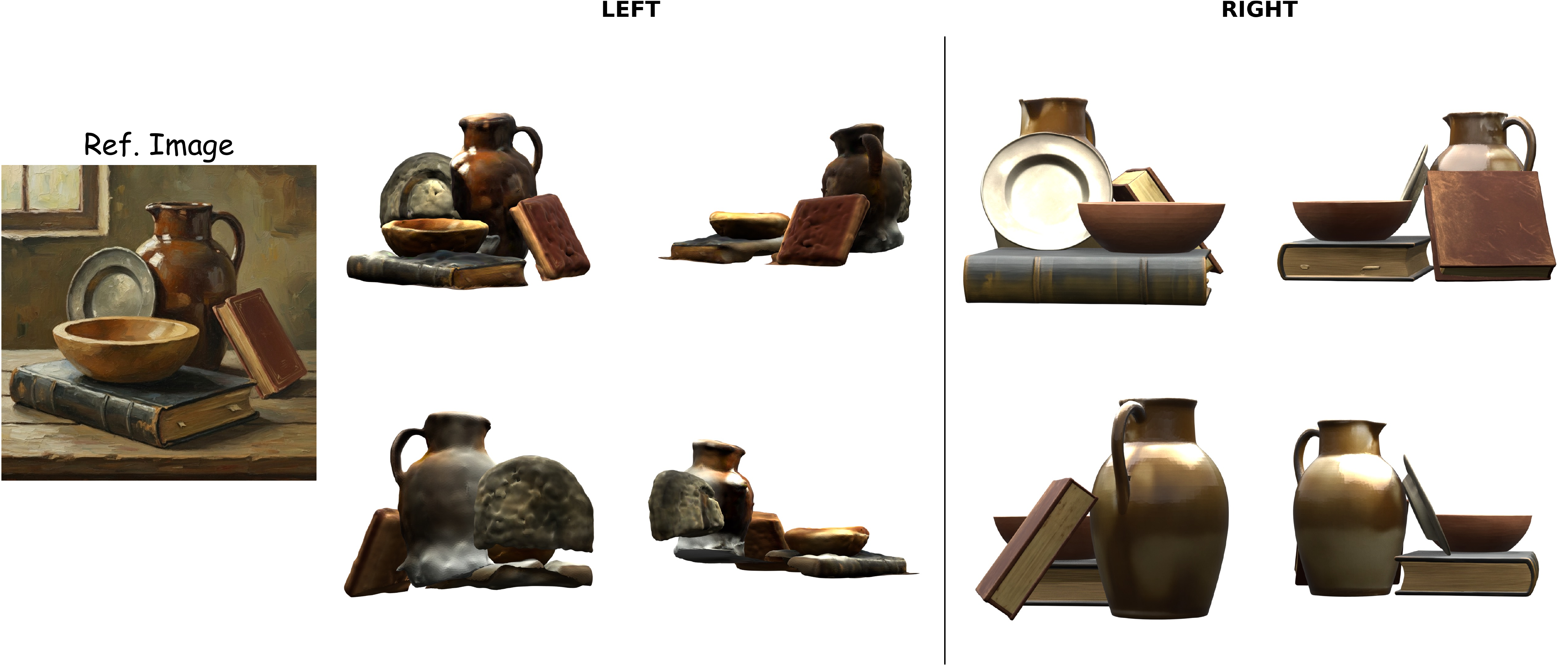}
    \vspace{-.6cm}
    \caption{\textbf{Left: Gen3DSR. Right: $\phi$-Scene.} In physical plausibility evaluation, $\phi$-Scene better preserves stable support relationships and reduces floating or interpenetrating objects across views compared to Gen3DSR.}
    \label{supp:fig:gen3dsr_vs_phi-scene}
\end{figure}

We also provide some example rendered scene pairs (Fig.~\ref{supp:fig:midi_vs_phi-scene}-\ref{supp:fig:gen3dsr_vs_phi-scene}) used in the MLLM evaluation, along with MLLM's outputs (Prompt~\ref{supp:prompt:midi_vs_phi-scene_visual}-\ref{supp:prompt:gen3dsr_vs_phi-scene_physical}) for each pair.

\begin{promptbox}[label={supp:prompt:midi_vs_phi-scene_visual}]{MIDI vs. $\phi$-Scene on Visual Quality}
\sloppy
Visual Quality:\\
Left: The left scene suffers from significant visual artifacts. The objects appear distorted and disconnected. The toolbox and bucket have strange, melted appearances, and the wooden crate has holes and inconsistencies. The wrench is also severely distorted and incomplete in the front view. The overall rendering is noisy and lacks clarity.\\

Right: The right scene presents much cleaner geometry and more recognizable object shapes. The surfaces and textures appear coherent, and the rendering is largely free from severe artifacts, holes, or distortions, especially when compared to the left scene. The object boundaries are clear, and the scene looks visually complete and consistent across the four views.\\

Comparison: The right scene is significantly better in terms of visual quality. It demonstrates much cleaner geometry, more accurate object shapes, coherent textures, and is largely free from the severe rendering artifacts and distortions that plague the left scene.\\

Final answer:\\
2
\end{promptbox}

\begin{promptbox}[label={supp:prompt:phi-scene_vs_sam3d_reference}]{$\phi$-Scene vs. SAM3D on Reference Alignment:}
\sloppy
Reference Alignment:\\
Left: The left scene includes all the main objects from the reference image. The shapes, proportions, and relative positions of the objects appear to be well-aligned with the reference. The red lid is positioned correctly and angled similarly. The colander's texture and position are accurate. The book "The Art of Baking" is present and in the correct location.\\

Right: The right scene also includes all the main objects. The shapes and proportions are generally good, but the positioning of some objects is slightly off. For example, the red lid seems a bit disconnected from the bowl it is meant to cover, and its angle is not as accurate as in the left scene. The colander's holes and texture are less distinct. The book's positioning also appears slightly different.\\

Comparison: The left scene is better because it more accurately represents the spatial layout and relative positions of the objects, particularly the red lid and the book. The shapes and textures of the objects, like the colander, are also rendered with greater fidelity to the reference image.\\

Final answer:\\
1
\end{promptbox}

\begin{promptbox}[label={supp:prompt:gen3dsr_vs_phi-scene_physical}]{Gen3DSR vs. $\phi$-Scene on Physical Plausibility}
\sloppy
Physical Plausibility:\\
Left: The left scene exhibits several issues. The books are stacked in a way that appears unstable, particularly the bottom book which is tilted. The jug is also positioned precariously on top of the books, with a significant portion of its base unsupported. The plate also seems to be floating or unrealistically supported. There is also significant interpenetration between objects, such as the books and the jug.\\

Right: The right scene shows a more physically plausible arrangement. The books are stacked more stably, and the jug is placed in a balanced position on top. The bowl is resting on the books, and the plate is resting on the jug, all of which appear to be stable configurations. There is minimal apparent interpenetration between objects.\\

Comparison: The right scene is significantly more physically plausible than the left scene. The arrangements in the left scene are unstable and violate basic principles of gravity and object support, with several instances of interpenetration. The right scene demonstrates stable stacking and placement of objects, making it appear much more realistic in terms of physical possibility.\\

Final answer:\\
2
\end{promptbox}

\section{UI for Human Evaluation}
\label{supp:sec:human}

\begin{figure}[H]
    \centering
    \includegraphics[width=\linewidth]{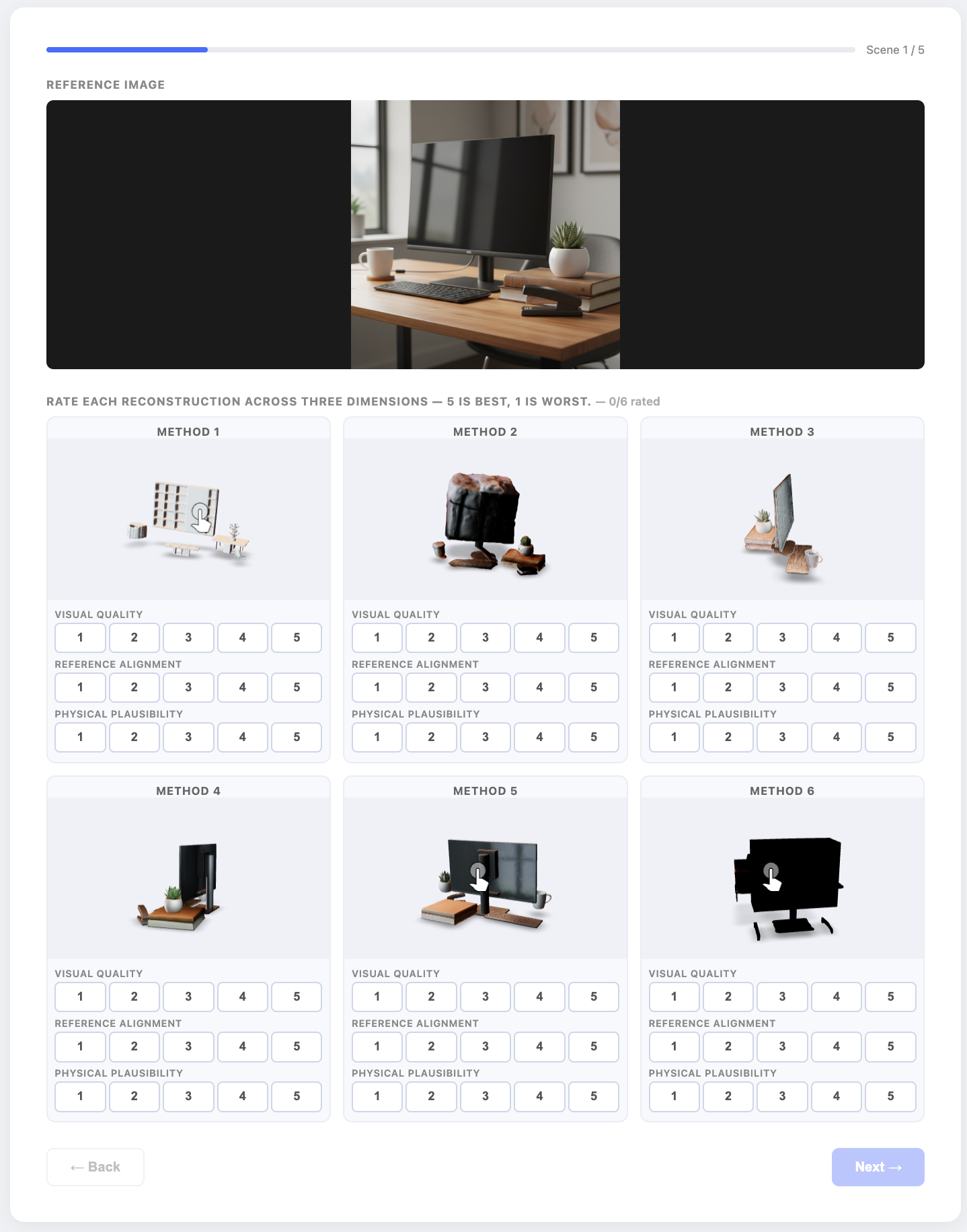}
    \vspace{-.6cm}
\caption{\textbf{Human evaluation webpage.}
The interface hides method names and allows participants to freely rotate, resize, and translate each reconstructed 3D scene for complete inspection.}
    \label{supp:fig:userstudy_screenshot}
\end{figure}

Following Sec.~\ref{sec:human_vlm_clip}, we provide a screenshot example (as illustrated in Fig.~\ref{supp:fig:userstudy_screenshot}) of the participant's webpage for the human evaluation results in Tab.~\ref{tab:human_vlm}.

\section{Failure Case Study}
\label{supp:sec:failurecase}
\subsection{Limitations of Collision-Proxy Approximation}
As shown in Fig.~\ref{supp:fig:failurecase}, $\phi$-Scene may still produce imperfect contacts, such as penetration and floating artifacts.
These artifacts mainly arise from the geometric mismatch between the visible raw meshes and the collision proxies used for simulation.
Although convex-decomposed compound colliders effectively improve rigid-body stability, they only approximate the reconstructed meshes and may not faithfully preserve fine local structures or concave regions.

This proxy-mesh mismatch leads to two complementary failure modes.
When the proxy under-approximates the raw mesh, i.e., the collision surface lies inside the visible surface or misses protruding/thin geometry, contact is detected too late.
As a result, two visible meshes may already overlap before their proxies collide, producing penetration, e.g., the green book leaning against the insulated bottle in the top-right of Fig.~\ref{supp:fig:failurecase}.
Conversely, when the proxy over-approximates the raw mesh, i.e., the collision surface lies outside the visible surface or becomes thicker than the actual geometry, contact is detected too early.
The object is then stopped before the visible surfaces touch, producing floating, e.g., the green book above the brown bowl in the bottom-right of Fig.~\ref{supp:fig:failurecase}.
These artifacts reflect the inherent trade-off between robust rigid-body simulation and exact geometric fidelity when using approximated collision proxies instead of raw mesh collision.

\subsection{\mbox{Limitations of Object Reconstruction}}
In addition to contact artifacts, $\phi$-Scene may inherit object-level errors from the utilized compositional 3D foundation model~\cite{chen2025sam}.
Since each object is reconstructed from partial and potentially occluded observations, its geometry may deviate from the reference object.
For example, in Fig.~\ref{supp:fig:failurecase}, the spoon is reconstructed too short and the green book becomes overly thick.

\subsection{\mbox{Limitations of Texture Refinement}}
\begin{figure}[t]
    \centering
    \includegraphics[width=\linewidth]{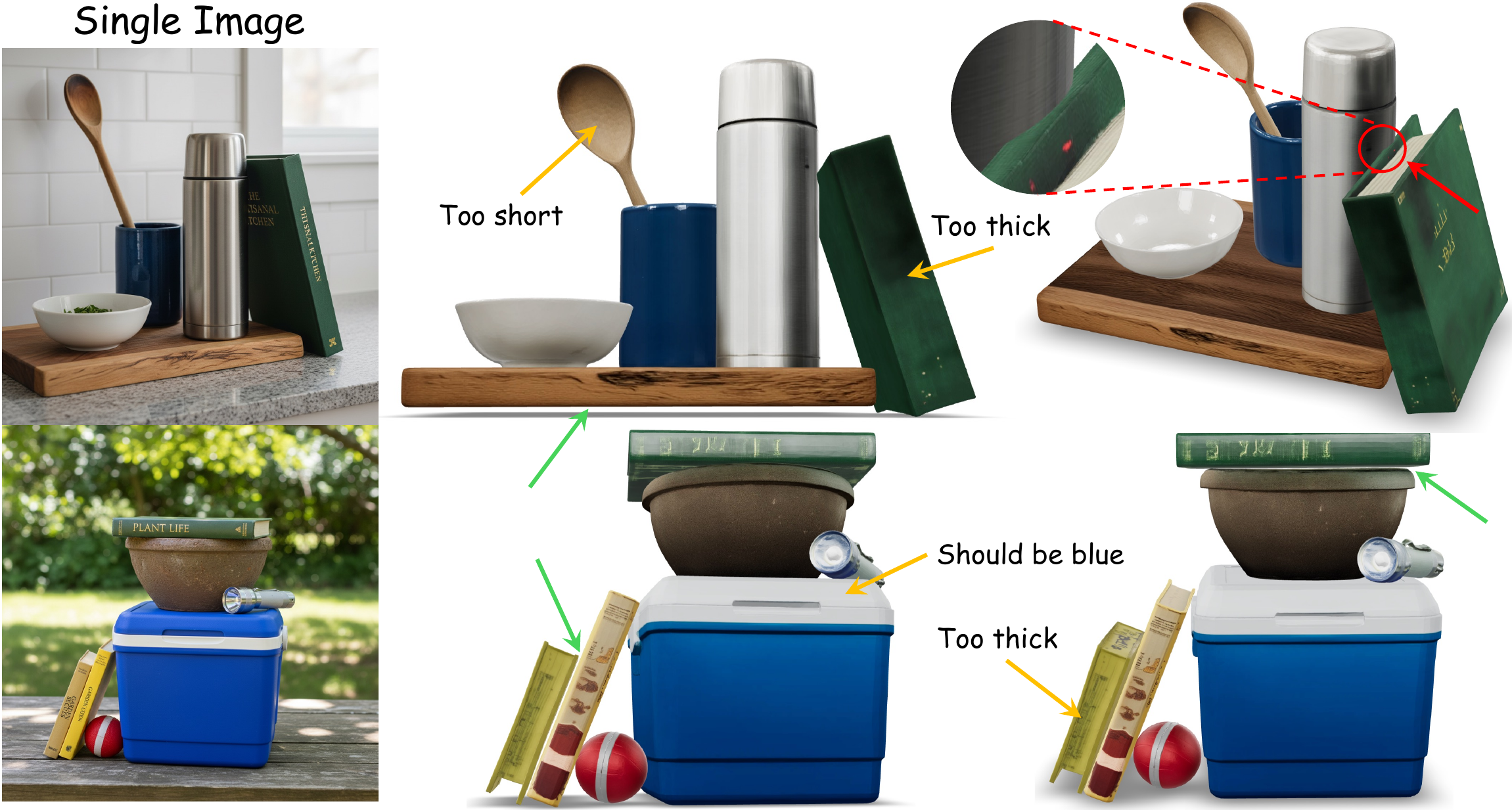}
    \vspace{-.6cm}
\caption{
\textbf{Failure cases.}
Although $\phi$-Scene substantially improves physical grounding and scene coherence, it can still produce residual artifacts.
The arrows highlight representative failure regions: red indicates penetration, green indicates floating (more visible when zoomed in), and yellow indicates inaccurate object geometry or texture.
}
    \label{supp:fig:failurecase}
    \vspace{-.3cm}
\end{figure}

Our object texture refinement stage can also introduce appearance deviations.
Although it generally improves texture sharpness and material realism, as shown in Supp.~Fig.~\ref{supp:fig:texrefine_cases}, the refined texture may drift from the reference image, e.g., the cooler lid should have a blue top but is reconstructed as white in Fig.~\ref{supp:fig:failurecase}.
These failures suggest that further improvements in object-level geometry reconstruction and reference-faithful texture refinement would complement our physical grounding module.

\clearpage

{
    \small
    \bibliographystyle{ieeenat_fullname}
    \bibliography{main}
}


\end{document}